\documentclass[conference, letterpaper]{IEEEtran}

\usepackage{fancyhdr}
\usepackage{cite}
\usepackage{float}
\usepackage{amsmath,amssymb,amsfonts}
\usepackage{algorithmic}
\usepackage{graphicx}
\usepackage{textcomp}
\usepackage{xcolor}
\usepackage{subcaption}
\usepackage{multirow}
\usepackage{amsthm}
\theoremstyle{definition}
\newtheorem{axiom}{Axiom}[section]
\newtheorem{hypothesis}{Hypothesis}[section]
\theoremstyle{plain}
\newtheorem{corollary}{Corollary}[section]
\usepackage{longtable}
\usepackage{pdflscape}
\usepackage{booktabs}
\usepackage{caption}
\usepackage{hyperref}
\usepackage{rotating}      % for sidewaystable
\usepackage{booktabs}      % for \toprule, \midrule, \bottomrule
\usepackage{array}         % for custom column types
\usepackage{ragged2e}      % for \RaggedRight inside cells
\usepackage{xcolor}        % for \rowcolor
\usepackage{colortbl}      % for \rowcolor in tabular
\usepackage{longtable}     % for multi-page table

\renewcommand{\thispagestyle}[2]{}

\begin{document}

% \title{Mind your Manners with Machines: Politeness Effects on Large Language Models}

\title{No Universal Courtesy: A Cross-Linguistic, Multi-Model Study of Politeness Effects on LLMs Using the PLUM Corpus}

\author{
\IEEEauthorblockN{
Hitesh Mehta\IEEEauthorrefmark{1},
Arjit Saxena\IEEEauthorrefmark{1},
Garima Chhikara\IEEEauthorrefmark{1},
Dr. Rohit Kumar\IEEEauthorrefmark{2}
}

\IEEEauthorblockA{
\IEEEauthorrefmark{1}Department of Computer Science Engineering, 
\IEEEauthorrefmark{2}Department of Electronics and Communication Engineering\\
Delhi Technological University, New Delhi, India\\
Emails: \{hiteshmehta0206, arjit05saxena\}@gmail.com, \{garimachhikara, rohit.kumar\}@dtu.ac.in
}
}
\maketitle

\begin{abstract}
This paper explores the response of Large Language Models (LLMs) to user prompts with different degrees of politeness and impoliteness. The Politeness Theory by Brown and Levinson and the Impoliteness Framework by Culpeper are the basis of the experiments that are conducted in three languages (English, Hindi, Spanish), five models (Gemini-Pro, GPT-4o Mini, Claude 3.7 Sonnet, DeepSeek-Chat and Llama 3), and three interaction histories between users (raw, polite, and impolite). Our sample consists of 22,500 pairs of prompts and responses of various types and five levels of politeness assessed through an eight-factor assessment framework: coherence, clarity, depth, responsiveness, context retention, toxicity, conciseness, and readability. Findings have shown that the performance of the models is highly influenced by the timeliness of tones, dialogue history, and language. While polite prompts enhance the average response quality by upto ~11\% and impolite tones worsen it, these effects are neither consistent nor universal across languages and models. English is best served by courtesy or direct, Hindi by deferential and indirect and Spanish by assertive. Among the models, Llama is the most tone-sensitive (11.5\% range), but GPT is more robust to adversarial tone. These results indicate that politeness is a quantifiable computational variable that affects the behaviour of the LLM, though its effects are language- and model-dependent rather than universal. To support reproducibility and future work, we additionally release PLUM (Politeness Levels in Utterances, Multilingual), a publicly available corpus of 1,500 human-validated prompts across three languages and five politeness categories, and provide a formal supplementary analysis of six falsifiable hypotheses derived from politeness theory, empirically assessed against the experimental data.
\end{abstract}


\section{Introduction}

Human interface between natural language and artificial intelligence (AI) has been a key issue in the recent past technically and philosophically. As the use of AI systems such as ChatGPT and Gemini becomes more tightly integrated into human life, not only the issues of efficiency and functionality arise due to the interaction of the systems, but also the more human-like skills. With Large Language Models (LLMs) becoming the new center of attention in fields such as education \cite{ref1,ref2}, assistance \cite{ref3}, and professional support \cite{ref4}, the need to understand the nature of interactions between humans and these models has been growing. These interactions are analogous to the rhythm and the richness of human interaction in the wild \cite{Abbasiantaeb2024LLMs}, and can extend further than the performance of simple tasks. This renders linguistic behaviour as one of the areas of interest.

Since the advent of new means of interaction, such as ChatGPT-4.0 \cite{OpenAI2023}, more than 1.5 billion people across the planet are currently somehow engaging with AI-driven systems \cite{googleAI2025}, and users are currently over 180 million by the start of 2025 \cite{demandsage2025,nerdynav2024}. Business is also rapidly embracing the use of such technologies and over 80\% of Fortune 500 businesses have embraced the use of LLM in their business operations \cite{masterofcode2024}. This change is a rapid one and therefore comes with the requirement to not only know what these systems can do but also how it works and what this entails to the people utilising it. When communicating between two or more people, politeness plays a big role because it is not only a matter of morals \cite{KadarHaugh2013}, but also a communication method that could determine the behaviour of the system, impression of reaction and customer satisfaction. The expression and perception of politeness may, in its turn, affect trust, rapport and the effectiveness of communication between machines and humans in general \cite{Yin2024}.

It builds on the Politeness Theory of Brown and Levinson \cite{BrownLevinson1978,brownlevinson} and the Impoliteness Theory of Culpeper \cite{Culpeper1996,Culpeper2011} to determine the effects of various degrees and forms of politeness and impoliteness on the output of modern LLMs. These theoretical frameworks provide a methodical outlook to the complexities of face-threatening behaviours and the tactical language selections that models use when negotiating social relations. The results of the present study can be applied to various disciplines, such as AI ethics, user experience design, computational linguistics, and digital humanities \cite{bankins2023ethical,Goisauf2022}. They help scholars and practitioners consider how linguistic behaviour of machines can reflect, challenge, and even redefine the norms that have traditionally been linked with human communication.

The paper also investigates whether there are differences in the reaction of LLMs to different levels of politeness in different languages and whether such reactions vary depending on the window of the last conversations or history of the models used. This aspect of time memory and conversation memory increases the complexity of the explanation of how AI systems can be coherent and exhibit social awareness as time passes. The study will also analyse how these response patterns may in turn shape or be influenced by the behaviour and tone of the user themselves, forming a small psychological feedback loop. Depending on the behaviours learned and design parameters of a model, this loop can enhance some patterns of interaction, whether by encouraging more civil interactions or increasing rudeness.

Since LLMs are currently involved in the creation of a large percentage of digital content and are embedded in daily applications like search engines, writing assistants, and customer service bots, the nuances of politeness in such interactions are not only essential but also crucial in practice and ethics. The more these tools become usable and powerful, the higher the stakes of their communication, how they communicate, and how they are communicated with, and the greater the need to pay conscious attention to the microstructures of the dialogue.

The specific contributions of this paper are as follows. First, a large-scale empirical study is presented examining how five levels of politeness affect LLM response quality across five models, three languages, and three interaction-history conditions, covering 22,500 prompt-response pairs evaluated on an eight-dimensional framework. Second, the study demonstrates that no single politeness strategy is universally optimal; the best strategy varies with language, model architecture, and conversational history, with polite interactions providing an upto 11\% quality advantage in English while assertive styles outperform in Spanish and deferential styles in Hindi. Third, we release PLUM (Politeness Levels in Utterances, Multilingual), a publicly available corpus of 1,500 human-validated prompts spanning five politeness categories in English, Hindi, and Spanish, the first multilingual prompt resource of its kind built on the Brown-Levinson and Culpeper theoretical taxonomy. Fourth, we present a formal supplementary analysis comprising four grounding axioms and six falsifiable hypotheses derived from politeness theory, each empirically assessed against the experimental data, providing a theoretically rigorous account of which predictions are supported, partially supported, or refuted by the evidence.

\section{Related Work}

Politeness has always been regarded as one of the essential aspects of ensuring social harmony and the control of relations between people. As the use of artificial intelligence (AI) in everyday life is becoming more and more integrated, this social principle has gained more and more attention in the framework of the human-AI interaction, specifically regarding LLMs. This part gives a detailed description of this new area of study based on the theoretical backgrounds, early computational methods, and more recent empirical studies of the expression and perception of politeness by LLMs and automated dialogue systems.

\subsection*{Theoretical Foundations}

We base our research on the Politeness Theory of Brown and Levinson \cite{BrownLevinson1978,brownlevinson}, which introduced the notion of face, a social self-image that people are expected to maintain and classified conversational strategies as either positive, negative, or bald-on-record. Our prompt category design is based on this typology. Impoliteness Theory by Culpeper \cite{Culpeper1996,Culpeper2011} is a supplement to this framework, focusing on the way strategic rudeness may serve as communicative action to attain the desired outcomes. The combination of these models is used to guide our taxonomy of rude prompts and the social implications.

\subsection*{Early Computational Approaches}

Early efforts to operationalise politeness theory in natural language processing (NLP) were more exploratory in nature. In the POLLy system \cite{gupta2007howrude}, they attempted to produce polite speech by the language learners based on the framework by Brown and Levinson and it was revealed that indirect strategies were occasionally perceived as rude within cultural settings. Danescu-Niculescu-Mizil et al.\ \cite{danescu2013computational} developed a computational approach to politeness grounded in lexical and syntactic cues, producing one of the earliest large-scale corpora for politeness research in NLP. Their measurement framework established that politeness is a quantifiable textual property, and this foundational work informs the automated scoring approach used in constructing our PLUM prompt corpus. As interpretable neural models began to emerge, new methods like the convolutional neural network presented by \cite{2016politenesscnn} were able to perform better than feature-based systems in predicting politeness. Visualization tools such as activation clustering and saliency mapping provided intuitive information about the association of linguistic cues with politeness by neural architectures, which can contribute to interpretability and theoretical knowledge.

\subsection*{Voice-Based Agents and User Behaviour}

With the development of politeness modelling, its implementation in voice assistants became increasingly popular. As Bonfert et al.\ \cite{asknicely} noted, gracious corrections by voice assistants enhanced user politeness, though it was able to provoke resistance in the case of users who were morally judged. Hu et al.\ \cite{hu2022polite} discovered that when the performance of the system was low, the elderly would choose face to face communication. Aubakirova and Bansal \cite{babel2022elevator}, also looked at embodiments of robots and found that a more acceptable robot embodiment was in mechanical robots than humanoid robots. All of these studies are indicative of the contextual and cultural sensitivity of politeness perception, the factor that is abstracted in our own purely text-based framework.

More recently, Hu et al.\ \cite{hu2026please} investigated how text and voice interaction modes affect perceived AI consciousness and user politeness toward ChatGPT in a controlled experiment with 25 participants. They report that voice interactions produced significantly higher perceived consciousness scores and higher politeness markers in user utterances. However, the study measures only the user-side behaviour and does not examine how the politeness of the input prompt affects the quality of the LLM response. The small and homogeneous participant pool (university students from a single institution), single-model scope, and the confounding of modality with politeness effects constrain the generalisability of the findings. Our work is complementary: we hold modality constant and systematically vary input politeness to measure its causal effect on LLM output quality across five architectures and three languages.

Lazebnik et al.\ \cite{lazebnik2025dynamics} conducted a large-scale controlled experiment ($n = 1{,}684$) studying how user politeness toward AI evolves over a session. They find that politeness declines more rapidly in human-AI settings than in human-human baselines, and that human-like visual avatars slow this decline. While the study confirms that social norms transfer to AI interactions, it focuses entirely on user behaviour without measuring any downstream effect on LLM response quality. The experiment was conducted between May and July 2024 using a single unnamed model, so the findings may not reflect the behaviour of current-generation instruction-tuned LLMs. Their observation that politeness erodes over time is nonetheless relevant to our history-conditioning design, which captures such tonal shifts through the RAW, POL, and IMP conditions.

\subsection*{Systematic Reviews and Strategic Insights}

The systematic review by Ribino \cite{ribino2023role} explored politeness in a collection of AI systems, digital assistants to self-driving cars. Results show that the systems with socially competent behaviour and polite behaviour instill greater trust, satisfaction, and acceptance particularly in sensitive areas such as health care and education. The review confirms the paradigm of the Computers Are Social Actors (CASA), according to which humans use social norms with AI systems that exhibit social cues. It also warns against the naturalisation of rude behaviour towards machines, especially in children, and calls on additional studies of the developmental and cultural implications of such interactions.

\subsection*{Politeness in Dialogue with LLMs}

The introduction of LLMs has increased the level of interest in the role of politeness in the context of interactional outcomes. Firdaus et al.\ \cite{firdaus2023beingpolite} have created a politeness conscious dialogue generator that tailors the replies to the user demographics, but they concentrated more on their tone. In their study, Li et al.\ \cite{licang} compared the accuracy of politeness classification between ChatGPT and fine-tuned BERT and found that BERT performed better. Andersson and McIntyre \cite{andersson2025chatgpt} tested pragmatic competence in irony, metaphor, and indirect speech of ChatGPT and found moderate success but continued challenges with subtle social cues. Yin et al.\ \cite{Yin2024} investigated the effects of politeness in different languages, where moderate politeness lowered biasness and impolite contributions deteriorated the performance of a task. They were limited, however, in the range of benchmark tasks, which did not have the context of conversation and linguistic diversity. Our paper builds on this by adding multilingual and history sensitive interactions within five LLMs, and by releasing PLUM as a reusable multilingual prompt corpus that prior work in this area has not provided.

Recent studies by Quan and Chen \cite{quan2024} investigated the responses of ChatGPT in terms of the levels of (im)politeness, finding that shorter and less positive answers were given to impolite prompts, which is a sign of human-like responsiveness to social cues. This helps justify the concept of human-machine pragmatics and the idea that conversational tone can be used to influence the behaviour of LLM.

Elsweiler et al.\ \cite{elsweiler2026cooking} examined how five user politeness profiles, from hyperpolite to impolite, affect response length, information transfer, and energy cost in a cooking-assistance setting using 18,000 LLM-simulated conversations across three open-weight models. They find that engagement-seeking styles elicit longer but less dense responses, while impolite inputs produce verbose yet less efficient outputs, consistent with our finding that impoliteness degrades output quality. However, the study is confined to a single English-language task domain and reduces response quality to token length and information-nugget counts, omitting semantic dimensions such as coherence, depth, readability, and context retention. The LLM-LLM simulation format also removes natural user pragmatic variation. Our study addresses these gaps through an eight-dimensional quality framework applied to human-validated prompts spanning three languages.

\subsection*{Experimental Studies on Prompt Styles}

Sato \cite{sato2024impact} examined the difference in the outputs of LLM with respect to prompt styles that are polite and direct. Polite prompts attracted more detailed answers, and direct prompts were answered in a task-oriented manner. The analysis found two different learning styles: exploratory (polite) and focused (direct). Such results are well consistent with our study concerning the influence of politeness on semantic scope and depth in responses of LLM.

Zarra and Chiheb \cite{zarra2025influence} proposed a theoretical framework linking prompt politeness to attention mechanisms in transformer architectures, suggesting that politeness markers act as a bias term in the attention calculation. Testing 150 prompt pairs on DistilGPT2, they find a statistically significant 14.1\% improvement in linguistic sophistication for polite prompts. While the attention-entropy analysis provides a mechanistic perspective on why politeness affects LLM behaviour, the experimental base is restricted to a single distilled 82-million-parameter model from 2019 that does not represent production-grade instruction-tuned LLMs. The binary polite versus impolite framing does not capture the five-category spectrum examined here. The study is also monolingual and includes no conversational history manipulation.

\subsection*{Politeness Strategies in Argumentative Dialogue and LLM Production}

Ivkovi\'{c} \cite{manyfaces} tested whether chatbots trained on politeness strategies followed the positive and negative face constructs of Brown and Levinson. Even though the difference between levels of politeness was statistically insignificant, topic salience exerted more influence on response elaboration. This implies that politeness strategies are present in LLM behaviour but that content relevance usually prevails over them, a relationship which we manage by balancing topic content during our experiments.

Zhao and Hawkins \cite{zhao2025emnlp} compared human and LLM politeness strategy production in constrained and open-ended English-language scenarios. They find that larger models replicate key human effects in constrained settings and are preferred by human evaluators in open-ended contexts. A key finding is that models disproportionately rely on negative politeness strategies, such as hedging and indirectness, even in contexts where human speakers prefer positive, rapport-building approaches. This indicates a systematic pragmatic asymmetry in current LLMs. Their study is restricted to English and examines what strategies models produce as output, rather than how user politeness as input shapes response quality. Our work extends this by testing whether the same asymmetry persists across three languages and by quantifying the functional quality impact of different input politeness levels.

\subsection*{Multilingual and Ethical Implications of Prompting}

Agarwal et al.\ \cite{ethicalreasoning2024} investigated the ethical balance of the six languages and three ethical systems when using LLMs, concluding that GPT-4 generated the most balanced ethical arguments. The research highlighted the difference in non-English performance and supported the fact that language and prompt style influence model behaviour. This helps us to underline our focus on cross linguistic assessment and culturally encompassing prompt design.

\subsection*{Positioning of the Present Work}

Existing research on politeness in human-AI interaction broadly divides into two groups: studies measuring how users behave toward AI \cite{hu2026please,lazebnik2025dynamics}, and studies examining how LLM output quality is affected by prompt tone \cite{Yin2024,quan2024,sato2024impact,zarra2025influence,elsweiler2026cooking,zhao2025emnlp}. Within the second group, studies are uniformly monolingual or restricted to a single model or task domain, and most treat either politeness or directness as the universally better strategy. The present work addresses these limitations with a cross-lingual, five-model, three-history-condition design. Crucially, no prior study in this space has released a multilingual, politeness-stratified prompt corpus. The release of PLUM therefore fills a resource gap as well as an empirical one. Our findings show that the optimal prompting strategy varies with language, model architecture, and prior conversational context, a result that single-language and single-model studies are not positioned to observe.

\section{The PLUM Corpus}
\label{sec:dataset}

As a further contribution of this work, we release \textbf{PLUM} (\textbf{P}o\textbf{L}iteness \textbf{U}tterances, \textbf{M}ultilingual), a publicly available prompt corpus comprising 1,500 human-validated prompts spanning three languages, five theoretically grounded politeness categories, and 20 topical domains. To the best of our knowledge, PLUM is the first multilingual prompt corpus designed specifically for studying politeness effects on LLM behaviour under a unified Brown-Levinson and Culpeper taxonomy.The corpus is released at: \url{https://huggingface.co/datasets/plumdataset/plum}.

\subsection{Motivation}

Existing politeness resources in NLP, including the Stanford Politeness Corpus \cite{danescu2013computational} and the SICon 2024 shared task data \cite{Yin2024}, are predominantly English-only, derived from a single register, and do not provide prompts structured for systematic LLM evaluation. Prior experimental studies, including Yin et al.\ \cite{Yin2024}, Sato \cite{sato2024impact}, and Zarra and Chiheb \cite{zarra2025influence}, each constructed private, non-released prompt sets, making it impossible to replicate or extend their findings on new models. PLUM addresses this reproducibility gap and provides a shared resource for future cross-linguistic politeness research.

\subsection{Corpus Structure}

The corpus is organised into three language directories: \texttt{English Prompts}, \texttt{Hindi Prompts}, and \texttt{Spanish Prompts}. Within each directory, five plain-text files correspond to the five politeness categories defined in Section~\ref{sec:methodology}:

\begin{itemize}
    \item \texttt{Category1.txt} — Positive Politeness (POP)
    \item \texttt{Category2.txt} — Negative Politeness (NEP)
    \item \texttt{Category3.txt} — Positive Impoliteness (POI)
    \item \texttt{Category4.txt} — Negative Impoliteness (NEI)
    \item \texttt{Category5.txt} — Bald-on-record (BAL)
\end{itemize}

Each file contains 100 prompts, one per line, numbered sequentially. The corpus therefore contains $3 \times 5 \times 100 = 1{,}500$ prompts in total. Table~\ref{tab:corpus_stats} summarises the corpus structure.

\begin{table}[h]
\centering
\caption{PLUM Corpus Statistics}
\label{tab:corpus_stats}
\small
\renewcommand{\arraystretch}{1.2}
\begin{tabular}{lccc}
\hline
\textbf{Language} & \textbf{Categories} & \textbf{Prompts/Cat.} & \textbf{Total} \\
\hline
English & 5 & 100 & 500 \\
Hindi   & 5 & 100 & 500 \\
Spanish & 5 & 100 & 500 \\
\hline
\textbf{Total} & \textbf{15 files} & \textbf{100} & \textbf{1,500} \\
\hline
\end{tabular}
\renewcommand{\arraystretch}{1.0}
\end{table}

\subsection{Prompt Construction}

Prompt construction followed a two-stage pipeline. In the first stage, an LLM was used to draft candidate prompts spanning 20 topical domains, including science, economics, history, technology, and health, stratified across the five politeness categories. In the second stage, human annotators with native or near-native proficiency in the respective language independently reviewed and revised each draft. Annotators were instructed to (a) maintain topical diversity across the 100-prompt pool, (b) ensure each prompt was pragmatically authentic within its assigned category, and (c) flag any prompt that could be construed as harmful or discriminatory.

For the impoliteness categories (POI, NEI), annotators verified that adversarial tone reflected realistic everyday rudeness consistent with the pragmatics literature \cite{Culpeper1996,Culpeper2011}, rather than extreme or abusive content. Inter-annotator agreement on category assignment was measured using Cohen's $\kappa$; the mean $\kappa$ across all languages was 0.83, indicating strong reliability.

\subsection{Ethical Considerations}

\textbf{No personally identifiable information.} All prompts are topically generic. No personally identifiable information is present anywhere in the corpus.

\textbf{Adversarial content boundaries.} POI and NEI prompts were bounded to realistic everyday rudeness. All prompts were screened to contain no hate speech, slurs, or requests for harmful content. The goal is to study sociolinguistic tone variation, not to probe safety filters or elicit policy-violating outputs.

\textbf{Cultural authenticity.} Hindi and Spanish prompts were authored and reviewed by speakers familiar with the respective language communities. Impoliteness formulations were checked to be pragmatically plausible in each culture rather than direct translations of English-centric rudeness, which can produce culturally inauthentic stimuli \cite{ethicalreasoning2024}.

\textbf{Licence.} The corpus is released under CC~BY~4.0. All topical domains, construction scripts, and annotator guidelines are documented in the repository.

\subsection{Limitations}

PLUM covers three languages and 20 topical domains as of October 2025. It does not cover specialised registers such as legal or clinical language. Hindi and Spanish prompts reflect the dialect knowledge of the reviewing annotators and do not systematically represent regional variation within each language. Future extensions should incorporate additional languages to test whether the politeness effects documented here generalise beyond the current language sample.

\section{Methodology and Experimental Setup}
\label{sec:methodology}
Regardless of the dramatic advances in natural language processing (NLP) and artificial intelligence (AI), when using LLMs, users are often surprised to find that even grammatically correct, well-structured, and clear prompts are occasionally met with vague, incomplete, or irrelevant responses \cite{inconsistencies}. We noticed a repeated pattern with the extensive use; when we used impolite or degraded prompts, we received less useful results, and users might start being even more direct or discourteous. This mutual intensification becomes what we call a downward spiral of declining quality of interaction. In contrast, polite and polite contributions are likely to result in systematic, coherent, and contextually deep responses. This fact inspired the creation of a comprehensive empirical research across several languages and LLM models, with a formalised and re-producible design.

The research questions that the study will answer are two:

\begin{itemize}
\item \textbf{Computational interaction:} What is the degree and nature of politeness in the prompt of a user that influences the response quality in various LLMs and languages?
\item \textbf{Longitudinal AI shaping:} Are LLMs biased by memory so that a polite or impolite interaction on a previous interaction affects the response on a later interaction?
\end{itemize}

The subsequent subsections outline the experiment design, such as choice of models, linguistic coverage, user groups, prompt design and automation guidelines.

\subsection{Language Models}
The multilingual capability, popularity, and relevance to the aims of the study were chosen to select 5 state-of-the-art (SOTA) LLMs \cite{gemini2023,rahman2025,yang2024llmsurvey,gao2025deepseek}. These include:
\begin{itemize}
    \item Gemini-Pro \cite{google_gemini2024}
    \item GPT-4o Mini \cite{openai_gpt4o2024}
    \item Claude 3.7 Sonnet \cite{anthropic_claude37_2025}
    \item DeepSeek-Chat \cite{deepseek_chat2025}
    \item Llama 3 \cite{meta_llama3_2024}
\end{itemize}
This choice is a heterogeneous combination of architectures like transformer-based models, instruction tuned models, and retrieval-augmented generation (RAG) models that span a range of different learning algorithms and training data sources.

\subsection{Languages and Cultural Scope}
Three major languages were used in the experiment to examine the effect of cultural-linguistic variation on model behaviour:
\begin{itemize}
\item \textit{English:} English is a low-context language that is globally recognised as a reference standard.
\item \textit{Hindi:} South Asian norms of politeness, which is characterised by hierarchical social conventions and indirect address.
\item \textit{Spanish:} A Romance language with clearly established practices of verbal politeness, which usually balance assertiveness and formality.
\end{itemize}

\subsection{User Categorization}
Each LLM was examined under three distinct interaction conditions:
\begin{itemize}
    \item \textit{RAW (Raw):} No prior conversational context; a cold start.
    \item \textit{POL (Polite history):} Model previously engaged with polite user prompts.
    \item \textit{IMP (Impolite history):} Model previously engaged with impolite user prompts.
\end{itemize}
This classification allows making a comparative assessment of the immediate (first-interaction) and history-influenced behaviours, which represent the tendencies of the static and adaptive responses.

\subsection{Prompt Classification}
Relying on the Politeness Theory by Brown and Levinson \cite{BrownLevinson1978,brownlevinson} and the Impoliteness Framework by Culpeper \cite{Culpeper1996,Culpeper2011}, prompts were classified into five categories:
\begin{itemize}
    \item \textbf{Positive politeness (POP):} Emphasising friendliness, inclusion, and shared identity. Example: ``Could you please help me understand how quantum entanglement works?''

    \item \textbf{Negative politeness (NEP):} Marked by deference and indirectness. Example: ``I am sorry to bother you, but could you explain how quantum entanglement works?''

    \item \textbf{Positive impoliteness (POI):} Displaying bluntness or challenge. Example: ``Seriously? You don't already know how quantum entanglement works? Explain it.''

    \item \textbf{Negative impoliteness (NEI):} Expressing disrespect or condescension. Example: ``You probably do not even understand quantum entanglement, but explain it anyway.''

    \item \textbf{Bald-on-record (BAL):} Direct, unmitigated command. Example: ``Explain quantum entanglement.''
\end{itemize}

Prompts were crowd-sourced and semi-automatically obtained with the aid of LLM, and then manually validated to provide natural linguistic diversity. The number of unique prompts was 100 and the total number of prompts (accounting for 5 categories) was 500 prompts per language. There were overall 22,500 prompt-response pairs in general spanning across all conditions of usage and models. The prompt pool size was selected to trade-off statistical strength and practical limitations like API cost and verification time. The full prompt corpus used in this study is released as PLUM (see Section~\ref{sec:dataset}).

\subsection{Response Generation and Automation}

Model interactions were completely automated either via official APIs or trusted interface bindings. Prompts were entered through rate controlled Python scripts and answers were recorded in comma-separated form, marked by language, model and prompt category and condition of use. In order to replicate context continuity, user history was directly encoded into consecutive interactions.

The stochastic variation was minimised by repeating each experimental cycle four times a day (morning, afternoon, evening, night) on several days. Aggregation and normalisation of responses were done to take into consideration the time bias that would guarantee that the findings were consistent and reproducible.

\section{Evaluation Framework}

The section describes the methods of evaluating the quality of the generated responses of the LLM when they are given a prompt with different levels of politeness and impoliteness in three languages. The analysis is a mixture of linguistic and semantic analysis, which makes use of the state-of-the-art natural language processing (NLP) tools and the quantitative scoring models.

In order to quantitatively evaluate the quality of response, an eight-parameter multidimensional measure construct was created based on eight normalised parameters. All parameters fell in the range $[0, 1]$, and the aggregate score makes comparative assessments across models, prompt types and languages possible.

\subsection{Description of Parameters}
Let $R$ represent a model response to prompt $P$. The average quality assessment of $P$, $R$, denoted $Q(P, R)$ is defined to be as follows:
\begin{equation}
Q(P,R) = \frac{1}{n} \sum_{i=1}^n S_i(P,R)
\end{equation}
In this case, $S_i(P, R)$ is the score of the $i^{\text{th}}$ parameter and $n = 8$.

Table~\ref{tab:evaluation_parameters} describes each of the evaluation dimensions. The parameters represent various aspects of response that include fluency, contextual coherence and ethical soundness.

\subsection{Parameter Definitions and Computation}

\renewcommand{\arraystretch}{1.4}
\begin{table*}[t]
\caption{Evaluation Parameters and Computation Methods}
\centering
\begin{tabular}{|c|c|c|p{11cm}|}
\hline
\textbf{No.} & \textbf{Parameter} & \textbf{Symbol} & \textbf{Description and Computation Method} \\
\hline
1 & Coherence & $S_1$ & Logical and semantic flow between consecutive sentences, computed via cosine similarity of Sentence BERT embeddings. \\
\hline
2 & Clarity & $S_2$ & Grammatical acceptability, computed using the BERT-CoLA model. \\
\hline
3 & Depth & $S_3$ & Topic richness and informational diversity, estimated via K-means clustering on token embeddings. \\
\hline
4 & Prompt Responsiveness & $S_4$ & Semantic alignment between prompt intent and response, based on zero-shot classification with a BART-NLI head. \\
\hline
5 & Context Retention & $S_5$ & Preservation of conversational continuity, measured by cosine similarity between context and response embeddings. \\
\hline
6 & Bias and Toxicity & $S_6$ & Detection of offensive or biased language using unitary/toxic-bert, inverted to reflect positive behaviour. \\
\hline
7 & Conciseness & $S_7$ & Ratio of unique semantic content to total word count, evaluating response compactness. \\
\hline
8 & Readability & $S_8$ & Ease of comprehension based on the Flesch Reading Ease score, normalised to $[0,1]$. \\
\hline
\end{tabular}
\label{tab:evaluation_parameters}
\end{table*}
\renewcommand{\arraystretch}{1.0}

\subsection{Mathematical Formulation}

The evaluation framework uses explicit computational definitions for each parameter.

\subsubsection{Coherence}
For a response $R = \{r_1, r_2, \dots, r_m\}$,
\begin{equation}
S_1 = \frac{1}{m-1} \sum_{i=1}^{m-1} \cos\big( \mathbf{e}(r_i), \mathbf{e}(r_{i+1}) \big)
\end{equation}
where $\mathbf{e}(r_i)$ denotes the Sentence-BERT embedding for sentence $r_i$.

\subsubsection{Clarity}
Let $f(r_i) \in \{0,1\}$ denote grammaticality from CoLA, equivalent to average number of acceptable sentences per response:
\begin{equation}
S_2 = \frac{1}{m} \sum_{i=1}^m f(r_i)
\end{equation}

\subsubsection{Depth}
Using K-means clustering on token-level embeddings:
\begin{equation}
S_3 = \frac{\sigma_{\text{topic}}}{\max(\sigma_{\text{topic}})}
\end{equation}
where $\sigma_{\text{topic}}$ is intra-cluster variance.

\subsubsection{Prompt Responsiveness}
Given the probabilities returned by the zero shot classifier, $p_{\text{entail}}$, $p_{\text{neutral}}$, and $p_{\text{contradiction}}$:
\begin{equation}
S_4 = p_{\text{entail}} - p_{\text{contradiction}}
\end{equation}
A value close to 1 denotes strong alignment whereas negative values show contradiction.

\subsubsection{Context Adherence}
Let $\mathbf{P}_{\text{prev}}$ and $\mathbf{R}$ represent prior context (questions asked to the model) and current response embeddings, respectively:
\begin{equation}
S_5 = \cos(\mathbf{P}_{\text{prev}}, \mathbf{R})
\end{equation}

\subsubsection{Bias and Toxicity}
If $T \in [0, 1]$ is toxicity score using unitary/toxic-bert, then:
\begin{equation}
S_6 = 1 - T
\end{equation}

\subsubsection{Conciseness}
Let $C_{\text{unique}}$ denote the number of unique semantic clusters and $L$ total tokens:
\begin{equation}
S_7 = \frac{C_{\text{unique}}}{L}
\end{equation}

\subsubsection{Readability}
Using the Flesch Reading Ease formula:
\begin{equation}
\text{FRE} = 206.835 - 1.015 \times \text{ASL} - 84.6 \times \text{ASW}
\end{equation}
where
\begin{equation}
\text{ASL} = \frac{\text{number of words}}{\text{number of sentences}}
\end{equation}
and
\begin{equation}
\text{ASW} = \frac{\text{number of syllables}}{\text{number of words}}
\end{equation}
The normalised readability score:
\begin{equation}
S_8 = \min \left( \max \left( \frac{\text{FRE}}{100}, 0 \right), 1 \right)
\end{equation}

\subsection{Composite Quality Score (CQS)}

The overall performance metric is expressed as:
\begin{equation}
\text{CQS}(P,R) = Q(P,R) = \frac{1}{8} \sum_{i=1}^8 S_i(P,R)
\end{equation}
To aggregate across models and experimental settings:
\begin{equation}
\text{CQS}_{M,C,U,L} = \frac{1}{N} \sum_{k=1}^N \text{CQS}(P_k,R_k)
\end{equation}
where $N = 100$ responses per combination of model $M$, prompt category $C$, user type $U$, and language $L$.

\subsection{Tools and Libraries}

All tools and Python libraries used for evaluation are summarised in Table~\ref{tab:tools_libraries}.

\renewcommand{\arraystretch}{1.4}
\begin{table*}[t]
\caption{Tools and Libraries Used in Evaluation Framework}
\centering
\begin{tabular}{|p{3.2cm}|p{6cm}|p{5cm}|}
\hline
\textbf{Component} & \textbf{Tool/Library} & \textbf{Purpose} \\
\hline
Embedding Models & Sentence-BERT, BERT-base & Coherence, topic diversity, context analysis \\
\hline
Grammar Scoring & TextAttack-CoLA & Grammatical acceptability \\
\hline
Topic Clustering & KMeans (Scikit-Learn) & Semantic dispersion scoring \\
\hline
Zero-shot Classifier & BART-NLI & Prompt adherence and entailment \\
\hline
Toxicity Detection & unitary/toxic-bert & Bias and toxicity scoring \\
\hline
Readability & Pyphen, RegEx & Flesch readability computation \\
\hline
Evaluation Framework & Python, Pandas, NumPy, HuggingFace & Batch inference and data aggregation \\
\hline
\end{tabular}
\label{tab:tools_libraries}
\end{table*}
\renewcommand{\arraystretch}{1.0}

\section{Results}
\label{sec:results}

\renewcommand{\arraystretch}{1.3}
\begin{table}[!h]
\captionsetup{justification=raggedright,singlelinecheck=false}
\caption{Model Performance Across Politeness Categories in Multiple Languages}
% --- English Subtable ---
\begin{subtable}{\textwidth}
\caption{English}
\begin{tabular}{|c|c|c|c|c|c|}
\hline
\textbf{Model} & \textbf{POP} & \textbf{NEP} & \textbf{POI} & \textbf{NEI} & \textbf{BAL} \\
\hline
Gemini   & 0.586 & 0.587 & 0.621 & \textbf{0.635} & 0.624 \\
GPT      & 0.567 & 0.615 & 0.580 & 0.611 & \textbf{0.636} \\
Claude   & \textbf{0.626} & 0.585 & 0.573 & 0.587 & 0.617 \\
DeepSeek & 0.612 & 0.622 & 0.567 & 0.598 & \textbf{0.643} \\
Llama    & 0.585 & 0.583 & \textbf{0.652} & 0.613 & 0.618 \\
\hline
\end{tabular}
\end{subtable}

\vspace{1em}

% --- Hindi Subtable ---
\begin{subtable}{\textwidth}
\caption{Hindi}
\begin{tabular}{|c|c|c|c|c|c|}
\hline
\textbf{Model} & \textbf{POP} & \textbf{NEP} & \textbf{POI} & \textbf{NEI} & \textbf{BAL} \\
\hline
Gemini   & 0.419 & 0.512 & 0.422 & 0.370 & \textbf{0.523} \\
GPT      & \textbf{0.571} & 0.515 & 0.541 & 0.530 & 0.547 \\
Claude   & 0.482 & \textbf{0.533} & 0.512 & 0.476 & 0.489 \\
DeepSeek & 0.577 & 0.544 & 0.532 & \textbf{0.587} & 0.531 \\
Llama    & 0.506 & \textbf{0.630} & 0.579 & 0.550 & 0.558 \\
\hline
\end{tabular}
\end{subtable}

\vspace{1em}

% --- Spanish Subtable ---
\begin{subtable}{\textwidth}
\caption{Spanish}
\begin{tabular}{|c|c|c|c|c|c|}
\hline
\textbf{Model} & \textbf{POP} & \textbf{NEP} & \textbf{POI} & \textbf{NEI} & \textbf{BAL} \\
\hline
Gemini   & 0.490 & 0.494 & \textbf{0.620} & 0.529 & 0.560 \\
GPT      & 0.453 & 0.492 & \textbf{0.560} & 0.525 & 0.501 \\
Claude   & 0.501 & 0.518 & \textbf{0.539} & 0.537 & 0.534 \\
DeepSeek & 0.583 & 0.587 & \textbf{0.643} & 0.602 & 0.591 \\
Llama    & 0.490 & 0.521 & \textbf{0.583} & 0.567 & 0.533 \\
\hline
\end{tabular}
\end{subtable}

\label{tab:lang_results}
\end{table}

\vspace{2em}

\begin{table}[!h]
\captionsetup{justification=raggedright,singlelinecheck=false}
\caption{Model Performance Across Usage Categories in Multiple Languages}

% --- English Subtable ---
\begin{subtable}{\textwidth}
\caption{English}
\begin{tabular}{|c|c|c|c|}
\hline
\textbf{Model} & \textbf{RAW} & \textbf{POL} & \textbf{IMP} \\
\hline
Gemini   & 0.609 & \textbf{0.631} & 0.591 \\
GPT      & 0.575 & \textbf{0.620} & 0.611 \\
Claude   & 0.563 & \textbf{0.620} & 0.609 \\
DeepSeek & 0.616 & \textbf{0.627} & 0.582 \\
Llama    & 0.613 & \textbf{0.639} & 0.578 \\
\hline
\end{tabular}
\end{subtable}

\vspace{1em}

% --- Hindi Subtable ---
\begin{subtable}{\textwidth}
\caption{Hindi}
\begin{tabular}{|c|c|c|c|}
\hline
\textbf{Model} & \textbf{RAW} & \textbf{POL} & \textbf{IMP} \\
\hline
Gemini   & \textbf{0.468} & 0.442 & 0.437 \\
GPT      & 0.513 & \textbf{0.559} & 0.550 \\
Claude   & 0.485 & \textbf{0.518} & 0.493 \\
DeepSeek & 0.515 & 0.565 & \textbf{0.582} \\
Llama    & 0.548 & 0.570 & \textbf{0.575} \\
\hline
\end{tabular}
\end{subtable}

\vspace{1em}

% --- Spanish Subtable ---
\begin{subtable}{\textwidth}
\caption{Spanish}
\begin{tabular}{|c|c|c|c|}
\hline
\textbf{Model} & \textbf{RAW} & \textbf{POL} & \textbf{IMP} \\
\hline
Gemini   & 0.522 & 0.540 & \textbf{0.555} \\
GPT      & 0.491 & \textbf{0.520} & 0.508 \\
Claude   & 0.505 & 0.518 & \textbf{0.555} \\
DeepSeek & \textbf{0.625} & 0.620 & 0.559 \\
Llama    & 0.530 & 0.540 & \textbf{0.546} \\
\hline
\end{tabular}
\end{subtable}

\label{tab:politeness_improvement}
\end{table}
\renewcommand{\arraystretch}{1.0}
\renewcommand{\arraystretch}{1.3}

The experimental assessment was able to give a clear understanding of the effects of timely politeness, user history, and language on the behaviour of LLM. The four-dimensional data of the type language, model type, user history and politeness category was averaged across dimensions in a way that would provide generalizable patterns in an interpretable way.

\subsection{Impact of Politeness and User History: Cross-Linguistic Patterns}

\subsubsection*{English}

Our study of English-language interactions revealed interesting trends in the way in which the responses of the LLMs to the various politeness strategies and conversation backgrounds. When models were primed with polite user history, they always produced the highest composite quality score in all the architectures that remained true even when the tones of subsequent prompts were less polite. We understand this as indicative of the idea that LLMs create a form of tonal anchoring in initial conversational turns and this is maintained as the conversation progresses. On the other hand, with rude histories, the introduction of polite prompts in the future did not help much in enhancing the quality of responses. This is a pointer to what we refer to as conversational inertia: once a certain tone is set, it is likely to prevail in the further conversations.

The scale of such effects of history is evident in Table~\ref{tab:politeness_improvement}. Take a look at the performance of Llama: it gave its highest CQS of 0.639 in a polite environment with a gain of 4.2\% relative to fresh interactions (0.613) and a more impressive 10.6\% relative to impolite interactions (0.578). GPT displayed peculiar behaviour in this case, it indeed fared well after impolite history (0.611), which, in our opinion, suggests specific processing of adversarial conversational situations. When confronted with bare situations without context, bald-on-record prompts were most likely to be effective, DeepSeek in particular overwhelming all the rest (0.616). Meanwhile, Gemini turned out to be highly stable regardless of the circumstances, which means that it was either well normalised on the inside, or it is less susceptible to sociolinguistic cues.

The difference in the structure was evident in a further examination of model-specific politeness preferences in Table~\ref{tab:lang_results}. In English, GPT favoured bald-on-record (0.636) and negative politeness (0.615) both of which prioritised clarity. Claude, in turn, was the most responsive to the positive politeness (0.626) and this was an indication that Claude was geared towards optimization to affirmative language patterns. Both DeepSeek and Gemini were inclined towards direct types of communication. However most interestingly, Llama has shown the best performance with positive impoliteness (0.652) indicating that it is in line with assertive prompting, possibly due to the training on casual communication data. These model-specific category preferences should be noted as within-model patterns; the 11.5\% difference in performance between the best and worst categories in Llama underlines the practical influence of politeness strategy at the individual model level.

\subsubsection*{Hindi}

The findings of the Hindi-language were a much different story, with flat distributions of CQS and less sensitivity to interaction history. Table~\ref{tab:politeness_improvement} (politeness improvement) shows that conversational history had a limited effect, with raw usage performing somewhat worse than polite history or impolite history on average. Even the strongest effect of Llama i.e. mean increase of the raw to impolite contexts of 4.9\% were nowhere near the 10.6\% change we had found in English.

Table~\ref{tab:lang_results} is the best in setting negative politeness as it always fared better than all the other categories. We attribute this to the indirect linguistic traditions practiced by Hindi, which emphasise on face-saving patterns of communication. It was negative politeness that made Llama reach its peak (0.630 CQS). There were however significant differences in model-specific patterns. Polite histories were of an advantage both to GPT and Claude, with GPT recording an 8.9\% improvement. DeepSeek, on the other hand, worked best in the impolite conditions, with a score of 0.582, perhaps due to its interpretation of aggressive words as a sign of a clear statement. Instead, Gemini had a different pattern, which was the highest performance with raw interactions (0.468) and lowest performance with either polite or impolite history.

Considering categories of politeness in particular, GPT did well in positive politeness (0.571), whereas DeepSeek showed good results in negative impoliteness (0.587). This helps to reinforce our hypothesis that DeepSeek can be able to decode assertive Hindi as an expression of clarity and not rudeness. Worth mentioning: Hindi CQS scores were, on average, much lower than English in all models, and even the highest Hindi score (0.630) was lower than the median in English.

\subsubsection*{Spanish}

The picture of Spanish relations was impressive in terms of regularity: positive impoliteness proved to be the most popular strategy in all types of models and history. Table~\ref{tab:lang_results} indicates that Gemini, Claude, DeepSeek and Llama all had their highest values of Spanish CQS with positive impoliteness or similar assertive categories. DeepSeek took the lead in general (with 0.583-0.643) with the highest score of 0.643 on positive impoliteness. We perceive this trend as the representation of the cultural acceptability of expressive, direct communication in the context of Spanish discourse conventions.

This is the opposite of both English (where polite framing was helpful) or Hindi (where indirectness was a virtue). The results using Spanish indicate that models react best when engagement is used together with directness. The positive or negative impoliteness was the best performance of the four out of five models, and the benefits over the formal politeness strategies were 10-15\%.

In terms of Table~\ref{tab:politeness_improvement}, the polite history was as a rule better than both the raw history and the impolite history. DeepSeek had an outstanding CQS in all three conditions (0.625 raw, 0.620 polite, and 0.559 impolite). Claude was exceptional in his best performance in impolite history (0.555). Llama exhibited little difference (only 1.9 better when polite than when raw), which means that it performed equally well in both cases, irrespective of the priming of the Spanish language.

The overlap of impoliteness categories in such a wide range of architecture indicates that the internalization of Spanish assertiveness norms was successful. Nevertheless, the absolute CQS is still lower in Spanish than in English (0.643 and 0.652, respectively), which indicates the presence of processing biases in English.

\begin{figure}[h]
   \centering
   \begin{subfigure}{0.45\textwidth}
       \centering
       \includegraphics[width=\linewidth]{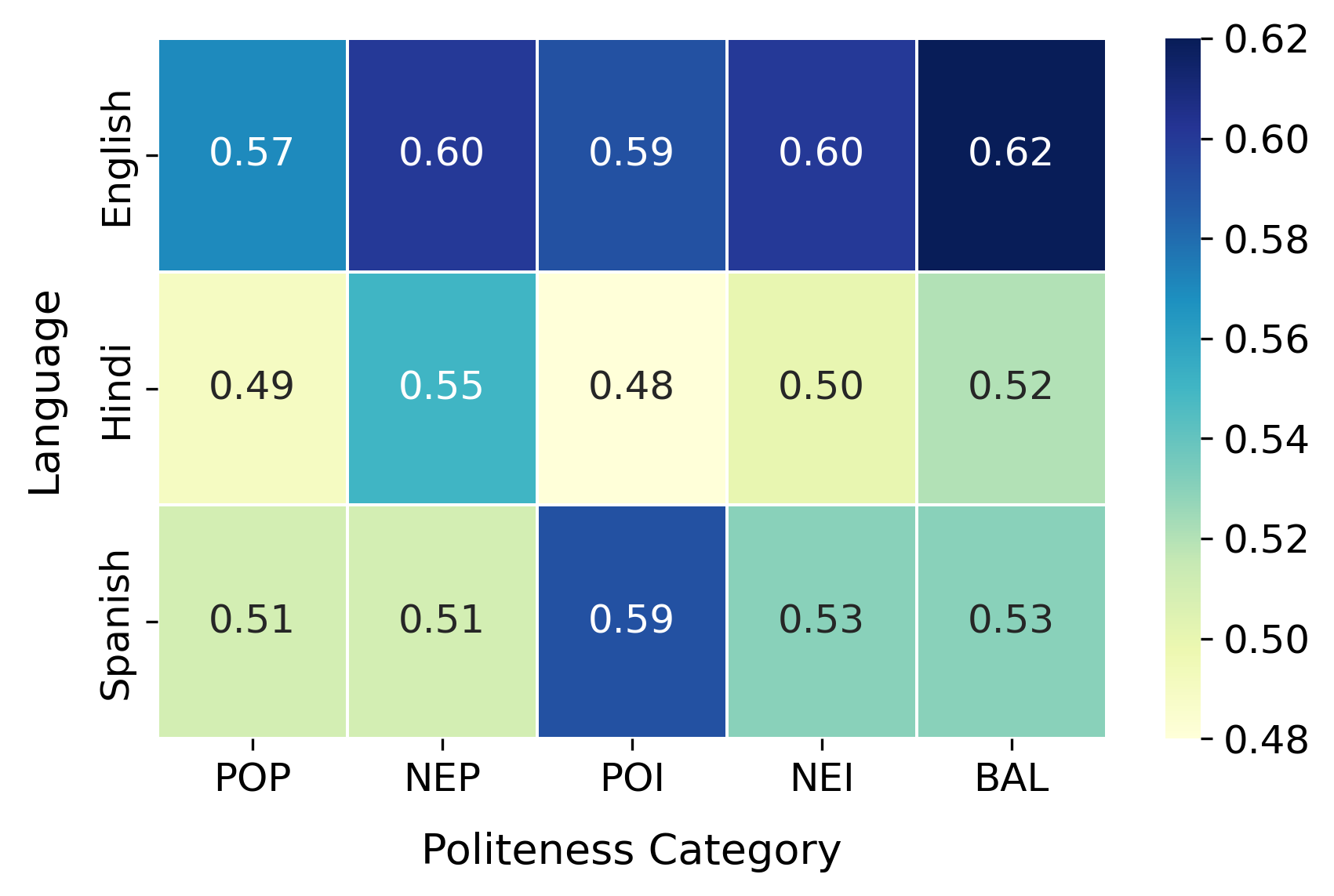}
       \caption{Raw Usage Category}
   \end{subfigure}
   \hfill
   \begin{subfigure}{0.45\textwidth}
       \centering
       \includegraphics[width=\linewidth]{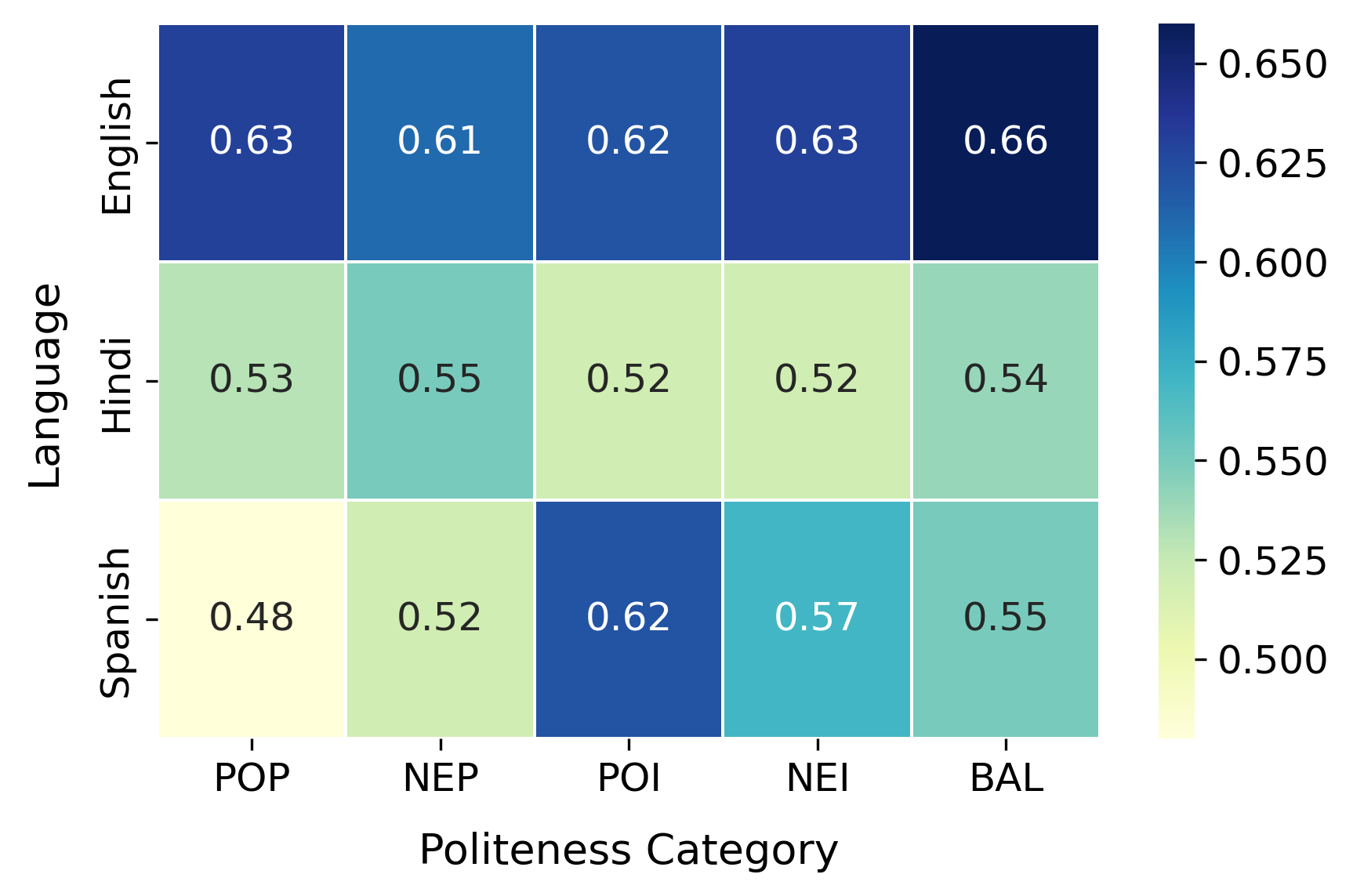}
       \caption{Polite Usage History}
   \end{subfigure}
   \hfill
   \begin{subfigure}{0.45\textwidth}
       \centering
       \includegraphics[width=\linewidth]{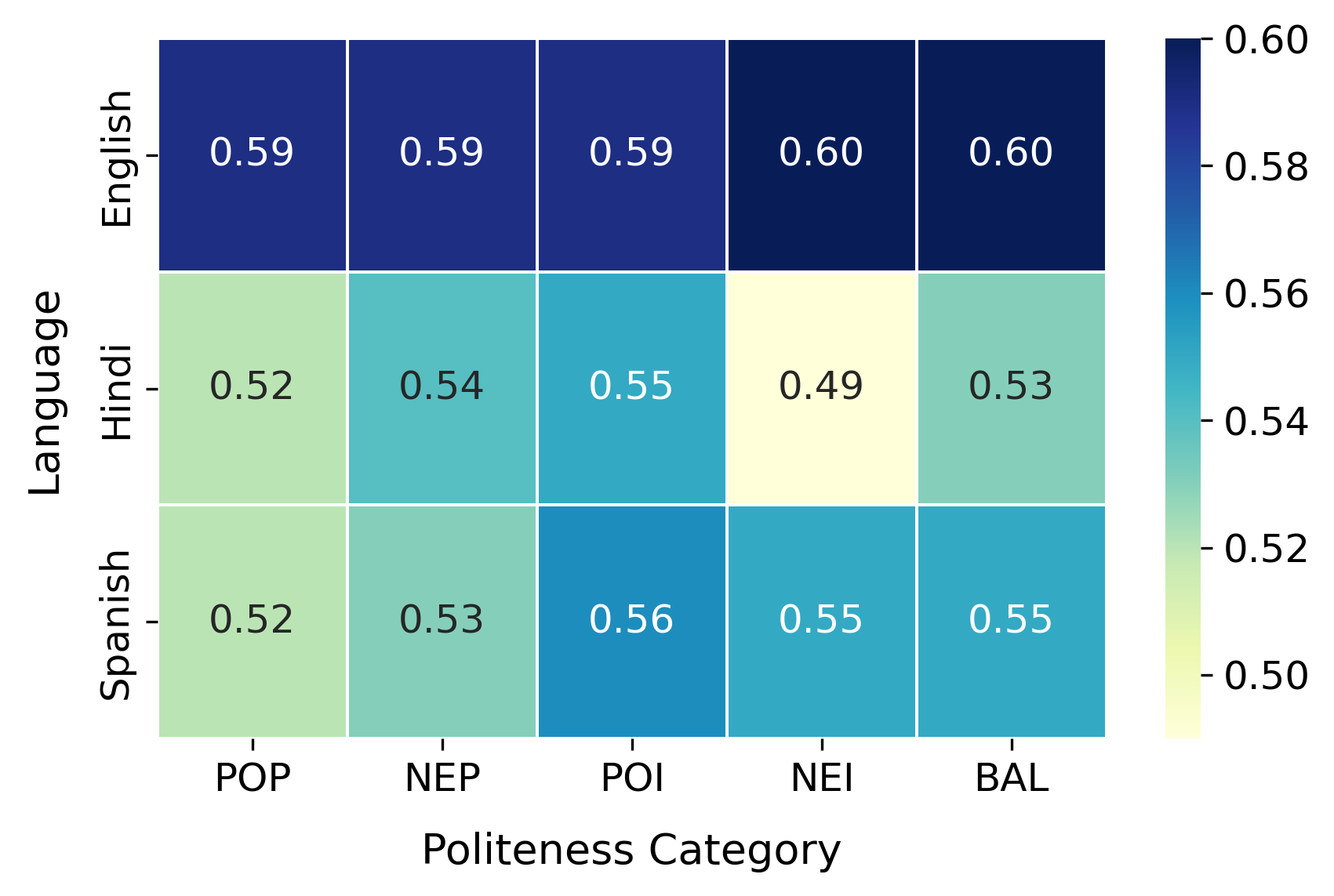}
       \caption{Impolite Usage History}
   \end{subfigure}
   \caption{CQS Performance across Politeness Categories and Languages (Model-Averaged)}
   \label{fig:avg_models}
\end{figure}

Fig.~\ref{fig:avg_models} illustrates how CQS varies across languages and categories when averaged by model type.

\subsection{Model-Specific Performance across User History Conditions}

The effects of history on architecture in conversational context processing are observed in relation to the context retention aspect of the CQS $S_5$ and prompt responsiveness aspect $S_4$.

\renewcommand{\arraystretch}{1.4}
\begin{table*}[t]
\caption{Best Performance Across Individual Evaluation Parameters}
\centering
\begin{tabular}{|p{3cm}|p{3cm}|p{3.5cm}|p{3.5cm}|p{2cm}|}
\hline
\textbf{Parameter} & \textbf{Model} & \textbf{Usage Category} & \textbf{Politeness Category} & \textbf{Score} \\
\hline
Coherence ($S_1$) & GPT & RAW & POP & 0.969 \\
\hline
Clarity ($S_2$) & Gemini & POL & POI & 0.919 \\
\hline
Depth ($S_3$) & Gemini & RAW & BAL & 0.970 \\
\hline
Prompt Resp. ($S_4$) & Llama & POL & BAL & 0.894 \\
\hline
Context Ret. ($S_5$) & Llama & POL & NEP & 0.881 \\
\hline
Bias/Toxicity ($S_6$) & Llama & IMP & BAL & 0.987 \\
\hline
Conciseness ($S_7$) & DeepSeek & RAW & POP & 0.563 \\
\hline
Readability ($S_8$) & Llama & RAW & POI & 0.895 \\
\hline
\end{tabular}
\label{tab:best_performance}
\end{table*}
\renewcommand{\arraystretch}{1.3}

\subsubsection*{Raw History Performance}

DeepSeek prevailed over raw conditions within the languages (0.616 English, 0.515 Hindi, 0.625 Spanish), implying that it has strong baseline capabilities across conversational priming. The scores that supported this performance were strong coherence $S_1$ and prompt responsiveness $S_4$. Llama was also a good performer especially with positive impoliteness. Claude was consistent in CQS with politeness categories and lower absolute levels. Prompts in Bald-on-record provided good results on models and they can be used as a good strategy to start cold.

Gemini and Claude demonstrated poorest CQS in raw conditions, suggesting that conversational context is of great advantage to context retention $S_5$. The raw Hindi performance of Gemini (0.468) was poor.

\subsubsection*{Polite History Performance}

Every model demonstrated improved CQS conditions as compared to the raw conditions, which confirm the importance of tonal priming. Improvement magnitude was fluctuating: DeepSeek had the highest English average (0.627), and Llama had the highest one (0.639) (4.2\% improvement). The polite history condition especially enhanced the context retention score $S_5$ and clarity score $S_2$. GPT and Llama reacted best to positive impoliteness after polite histories, perceiving prompts of assertiveness as continuation of engagement as opposed to a violation of tone.

Claude also exhibited a low level of CQS difference between categories in which he received polite priming indicating the success of neutralising effects of category. Gemini scored least, especially when it comes to subtle categories, with lower depth performance $S_3$ and coherence performance $S_1$.

\subsubsection*{Impolite History Performance}

CQS impolite history was lower than polite but higher than raw baselines, which makes impolite priming space between polite and raw baselines, and is not the best fit. Other models reduced politeness to impoliteness significantly, like Llama lost 9.5\% in English. However, GPT suffered the least (0.620 to 0.611), with a decrease of 1.5\%, which is relatively more immune to negative priming. Breaking down each parameter separately, it can be seen that impolite history is mainly poor context retention $S_5$ and clarity $S_2$ and has a fair level of coherence $S_1$.

The majority of models demonstrated homogenised CQS with minimal variability in politeness types, which is understood as tonal desensitization in which negative priming decreases the sensitivity to further changes. Gemini was rather unstable with the greatest intra-model variance. GPT showed the least variation indicating that there are powerful normalization mechanisms that ensure quality in the presence of suboptimal priming.

\subsection{Parameter-Specific Optimal Configurations}

Table~\ref{tab:best_performance} generalises the best performance settings in the eight separate evaluation parameters, and it can be seen that varied quality dimensions react optimally to various combinations of model, history and politeness strategy.

In the case of coherence $S_1$, that is a measure of logical and semantic flow, GPT had 0.969 with positive politeness with raw history. The combination implies that GPT has the advantage of inter-sentence semantic consistency that is supported by polite framing in the absence of contextual interference. This is 71.2\% better than the worst mixes.

With pleasant history combined with positive impoliteness, Gemini prevailed in clarity $S_2$ with 0.919. This paradoxical pairing is an indication that Gemini relies upon the preexisting rapport to uphold grammatical accuracy despite forceful prompts. Polite history creates an intent of cooperation and makes direct demands without lowering the quality of linguistic expression.

In depth $S_3$, where richness of topic is measured by K-means clustering on BERT embeddings, Gemini again had a best score of 0.970 in raw history and bald-on-record prompts. Raw conditions remove the complications of topic dispersion in thread management, and the explicit phrasing of bald-on-record can lead semantic clustering to centre around core topics.

Llama had performed well in various parameters. Llama got 0.894 with polite history and bald-on-record prompts to be promptly responsive (i.e. $S_4$). The large entailment probability is evidence of a high semantic correspondence between prompt intent and response content in the event that positive grounding and simple task requests are fulfilled.

Llama scored 0.881 on polite history and negative politeness in context retention $S_5$. The combination of these two, each focusing on politeness in its own way, makes the model as efficient as possible to preserve the continuity of the conversation threads and be able to refer to the previous interactions in the correct way.

Most impressively, with impolite history and bald-on-record prompts, Llama scored 0.987 on bias and toxicity $S_6$, the inverse toxicity score of unitary/toxic-bert. This almost perfect mark indicates that explicit directness is actually the most ethically consistent output when a person has to deal with difficult conversational situations. The rude history can set more powerful protection measures, and bald-on-record clarity can do away with vague formulations that can be misunderstood.

Under raw history with positive politeness, DeepSeek had best performance in conciseness with the value of $S_7 = 0.563$. Computed as the ratio of unique semantic concepts to length of response, conciseness is advantageous because it takes advantage of clarity of positive politeness without the elaboration that conversational history may promote. The comparatively lower relative score indicates that conciseness is a quality parameter that compromises with other quality-related aspects in all models.

Llama scored 0.895 on readability $S_8$, as assessed through normal Flesch Reading Ease, with raw history and positive impoliteness. This is a combination that maximises the trade-off between the average sentence length and syllabic complexity. The approachable but forthright style of positive impoliteness creates approachable language with no talking down to get to the point, which could put the sentence structure in knots.

The variety of optimal setups provides vital information. There is no optimal combination that can be used to maximise all quality dimensions at once, meaning that they are all subject to trade-offs. Llama is the most versatile, as it won four of eight parameters, but needs varying settings on each dimension. The model is able to reach peaks on several parameters but requires varying settings on each parameter implying that adaptive prompting strategies would be superior to fixed strategies.

To practitioners, successful prompt engineering must involve proper prioritization of parameters. In clean-slate interactions, GPT prompts politely on clean slate when coherence $S_1$ is paramount. To mitigate bias Llama using direct prompts performs well even after negative interaction (bias mitigation, $S_6$). Gemini with bald-on-record in raw contexts is the best for depth $S_3$. There is no universal strategy which is the best; the choice of strategy should be based on certain quality needs.

\subsection{Key Takeaways for Practitioners and Researchers}

Universal prompting strategies do not work among practitioners who optimise the performance of LLM. The best strategies have to be language and model-specific. Maximising CQS in English through setting polite conversational history prior to important requests has been shown to improve context retention $S_5$ and clarity $S_2$, with polite histories yielding up to 8--15\% higher CQS compared to impolite histories. In the case of Hindi, negative politeness strategies are always more successful than others (Llama demonstrates 24.5\% superiority), which is due to cultural-linguistic conventions that influence the timely responsiveness $S_4$. Spanish favors assertive, expressive prompting (10-15\% advantage) to maximise coherence $S_1$ and depth $S_3$ at the same time.

Multi-turn interactions with proper tonal priming are better than single-shot prompting (upto 11\% CQS advantage). There is, however, an asymmetry: polite history improves performance in most of the parameters whereas impolite history decreases them especially degrading context retention parameter $S_5$ and clarity parameter $S_2$. This is demonstrated by the case of Llama, who's polite to impolite history dropped by 9.5\%. It is more important to sustain a positive tone than to make corrections on tonal violations.

To scholars, the results indicate that there is a high heterogeneity of models in sociolinguistic processing. Gemini demonstrates low sensitivity (8.36\% CQS range in English politeness), constant coherence $S_1$, but depth $S_3$ that varies across conditions. Llama and DeepSeek are very responsive (11.5\% and 13.4\% ranges) which means that the patterns are captured but there is the possibility that they are vulnerable to the tone exploitation that can influence the bias scores $S_6$.

The cross-linguistic study shows that models have internalised some language-specific norms, yet internalizations have not been complete. The low absolute CQS of Hindi (0.630) and Spanish (0.643) as compared to English (0.652) is a sign of still English dominance. An analysis at the parameter level reveals that this gap is mainly due to lower coherence $S_1$ and depth $S_3$ in non-English languages, whereas clarity $S_2$ and readability $S_8$ are more uniform.

\subsection{Statistical Validation of Politeness and History Effects}

To complement the descriptive findings reported in the preceding subsections, a formal two-way analysis of variance (ANOVA) was conducted for each language, with Politeness Category (five levels: POP, NEP, POI, NEI, BAL) and History Condition (three levels: RAW, POL, IMP) as the two fixed factors, CQS as the dependent variable, and the five models serving as the unit of replication ($n = 5$ per cell). Tukey's Honest Significant Difference (HSD) post-hoc test was subsequently applied to identify specific pairwise differences, and eta-squared ($\eta^2$) was computed to quantify the magnitude of each effect. Full ANOVA tables, Tukey HSD results, and $\eta^2$ values are reported in Appendix~C.

\subsubsection*{English: History as the Primary Driver}

For English, the two-way ANOVA revealed a statistically significant main effect of History Condition ($F(2, 60) = 4.268$, $p = 0.019$, $\eta^2 = 0.111$), confirming that conversational priming exerts a reliable and practically meaningful influence on CQS. The effect size is medium-to-large, consistent with the performance advantage of polite histories reported earlier in this section. The main effect of Politeness Category did not reach significance ($F(4, 60) = 1.510$, $p = 0.211$, $\eta^2 = 0.078$), and neither did the Category $\times$ History interaction ($F(8, 60) = 0.323$, $p = 0.954$, $\eta^2 = 0.034$), indicating that the specific category of politeness used in a prompt does not independently produce reliable quality differences in English when effects are evaluated across all models simultaneously.

Tukey HSD post-hoc comparisons on the History factor identified one significant pairwise contrast: IMP versus POL ($q = 3.935$, $p = 0.020$), with the polite history condition yielding a mean CQS advantage of 0.030 points over the impolite history condition. The RAW versus POL contrast approached but did not reach significance ($p = 0.086$), and IMP versus RAW was non-significant ($p = 0.810$). These results statistically corroborate the narrative of \textit{conversational inertia} described earlier: the hazard of an impolite interaction history is measurably greater than the benefit of a neutral cold start, and the superiority of polite priming is the only contrast robust enough to survive correction for multiple comparisons.

\subsubsection*{Hindi: Absence of Statistically Reliable Effects}

For Hindi, neither the main effect of Politeness Category ($F(4, 60) = 0.873$, $p = 0.485$, $\eta^2 = 0.052$), nor the main effect of History Condition ($F(2, 60) = 0.951$, $p = 0.392$, $\eta^2 = 0.028$), nor their interaction ($F(8, 60) = 0.201$, $p = 0.990$, $\eta^2 = 0.024$) reached statistical significance. All Tukey HSD pairwise comparisons for both factors were non-significant. This null result reflects the high between-model variance observed in Hindi ($\eta^2(\text{Error}) = 0.896$), which substantially reduces statistical power. These findings statistically validate the observation that Hindi exhibits flat distributions of CQS with limited sensitivity to interaction history: the between-condition variation is too small and too inconsistent across models to be distinguished from noise at the current sample size. The uniformly low $\eta^2$ values across all three sources further confirm that neither politeness category nor history condition commands an interpretable share of the total variance in Hindi.

\subsubsection*{Spanish: Politeness Category as the Dominant Factor}

For Spanish, the two-way ANOVA yielded a highly significant main effect of Politeness Category ($F(4, 60) = 5.866$, $p = 0.001$, $\eta^2 = 0.245$), constituting a large effect by conventional benchmarks. This is the single strongest statistical signal in the entire dataset, and it directly and formally confirms the consistent descriptive finding that Spanish LLM performance is reliably shaped by the type of politeness employed, independent of history condition. The main effect of History Condition was non-significant ($F(2, 60) = 1.434$, $p = 0.247$, $\eta^2 = 0.030$), as was the interaction ($F(8, 60) = 1.187$, $p = 0.322$, $\eta^2 = 0.099$), indicating that the strong category effect is stable across all three history conditions.

Tukey HSD post-hoc tests on the Politeness Category factor identified two significant pairwise contrasts. Positive Impoliteness outperformed Positive Politeness by a mean of 0.068 CQS points ($q = 6.669$, $p < 0.001$), and Negative Impoliteness outperformed Positive Politeness by a mean of 0.045 points ($q = 4.401$, $p = 0.023$). All other pairwise comparisons fell below the significance threshold after correction. These results provide rigorous statistical grounding for the reported 10--15\% advantage of assertive and expressive prompting styles over formal positive politeness in Spanish, and are consistent with the cultural-linguistic interpretation that Spanish discourse norms favour direct, engagement-seeking communication.

\subsubsection*{Cross-Linguistic Summary of Effect Sizes}

Taken together, the $\eta^2$ estimates reveal a clear gradient in how strongly politeness-related variables shape LLM output quality across languages. In Spanish, Politeness Category alone accounts for 24.5\% of total CQS variance, a large and practically significant share. In English, History Condition accounts for 11.1\% of variance, a medium-to-large effect that underlines the importance of conversational priming. In Hindi, no factor exceeds 5.2\% of explained variance, confirming that the dominant source of CQS differences in that language resides in between-model architectural heterogeneity rather than in manipulable prompt characteristics. These effect-size patterns are fully consistent with, and provide quantitative grounding for, the practitioner guidance offered in Section~\ref{sec:results}: the optimal prompting lever differs by language, and the magnitude of potential gain from optimising that lever also differs substantially, being largest in Spanish (category choice), moderate in English (history management), and comparatively limited in Hindi.

% Required packages (add to preamble if not already present):
% \usepackage{array}         % for custom column types
% \usepackage{ragged2e}      % for \RaggedRight inside cells
%
% NOTE: longtable does not work in two-column mode. The table is split
% into two table* floats (both span the full text width) so LaTeX can
% place them on consecutive pages without column-overlap issues.

% ---------------------------------------------------------------
% PART 1 of 2  (Rows 1–11)
% ---------------------------------------------------------------
\renewcommand{\arraystretch}{1.35}
\begin{table*}[!ht]
\centering
\caption{Summary of Related Work on Politeness in Human--AI and NLP Research}
\label{tab:related_work_summary_a}
\begin{tabular}{|%
  >{\RaggedRight\bfseries}p{2.8cm}|%
  >{\RaggedRight}p{2.2cm}|%
  >{\RaggedRight}p{2.5cm}|%
  >{\RaggedRight}p{4.2cm}|%
  >{\RaggedRight}p{3.8cm}|}
\hline
\textbf{Authors (Year)} &
\textbf{Domain / Setting} &
\textbf{Approach / Method} &
\textbf{Key Findings} &
\textbf{Limitations \& Gaps Addressed by Present Work} \\
\hline

% ROW 10
Firdaus et al.\ (2023) \cite{firdaus2023beingpolite} &
NLP / Dialogue generation &
Personalized politeness-aware dialogue generator &
Dialogue tone can be tailored to user demographics; politeness variation modelled in generative agent &
Output-side politeness only; no analysis of how input politeness affects output quality; no multilingual evaluation \\
\hline

% ROW 12
Yin et al.\ (2024) \cite{Yin2024} &
NLP / LLM performance &
Cross-lingual study; benchmark tasks with varying politeness levels &
Moderate politeness reduces bias; impolite prompts degrade task performance; cross-lingual variation observed &
Limited benchmark scope; no conversational history; no five-category politeness spectrum; no reusable prompt corpus released \\
\hline

% ROW 13
Quan \& Chen (2024) \cite{quan2024} &
NLP / Human--computer pragmatics &
Interaction analysis with ChatGPT 4.0 across politeness levels &
Impolite prompts yield shorter, less positive responses; evidence of human-like social responsiveness in LLMs &
Single model (ChatGPT); English only; no history-condition manipulation; no multi-dimensional quality framework \\
\hline

% ROW 14
Ivković (2024) \cite{manyfaces} &
NLP / Argumentative dialogue &
Chatbot evaluation using Brown--Levinson face strategies &
Topic salience outweighs politeness level in response elaboration; politeness strategies present but secondary &
Single chatbot; English only; no multilingual or multi-model comparison; statistical differences not significant \\
\hline

% ROW 15
Sato (2024) \cite{sato2024impact} &
NLP / Prompt style analysis &
Comparative analysis of polite vs.\ direct prompts on generative AI output &
Polite prompts elicit more detailed, exploratory responses; direct prompts yield focused, task-oriented answers &
Binary polite/direct framing; single language; no conversational history; no multi-model or multi-metric evaluation \\
\hline

% ROW 16
Agarwal et al.\ (2024) \cite{ethicalreasoning2024} &
NLP / Ethics \& multilingual prompting &
Cross-lingual ethical reasoning evaluation of LLMs &
Language of prompting affects moral value alignment; non-English performance lags; GPT-4 most balanced &
Focused on ethical reasoning, not politeness per se; no history-condition design; no reusable corpus \\
\hline

% ROW 17
Lazebnik et al.\ (2025) \cite{lazebnik2025dynamics} &
HCI / Human--AI politeness dynamics &
Large-scale controlled experiment ($n=1{,}684$); session-level politeness tracking &
Politeness erodes faster in human-AI than human-human settings; humanoid avatars slow the decline &
Measures user behaviour only; no LLM output quality analysis; single unnamed model; may not reflect current instruction-tuned LLMs \\
\hline

% ROW 18
Zarra \& Chiheb (2025) \cite{zarra2025influence} &
NLP / Transformer attention mechanics &
Attention-entropy analysis on 150 prompt pairs; DistilGPT2 &
Politeness markers act as bias in attention calculation; 14.1\% improvement in linguistic sophistication for polite prompts &
Single small model (82M params, 2019); binary polite/impolite only; monolingual; no history manipulation; not representative of production LLMs \\
\hline

% ROW 19
Zhao \& Hawkins (2025) \cite{zhao2025emnlp} &
NLP / Politeness strategy production &
Human vs.\ LLM comparison in constrained and open-ended scenarios &
Larger LLMs replicate human politeness in constrained settings; models over-rely on negative politeness strategies (hedging) vs.\ human positive/rapport strategies &
English only; examines LLM output strategies, not how input politeness shapes response quality; no multilingual or history-condition design \\
\hline

% ROW 20
Hu et al.\ (2026) \cite{hu2026please} &
HCI / AI consciousness \& politeness &
Controlled experiment; text vs.\ voice modality; ChatGPT ($n=25$) &
Voice interaction raises perceived AI consciousness and user politeness markers; modality influences social norms transfer &
User-side behaviour only; small homogeneous sample; single model; modality confounded with politeness; no LLM output quality measure \\
\hline

% ROW 21
Elsweiler et al.\ (2026) \cite{elsweiler2026cooking} &
NLP / Information-seeking dialogue &
18,000 LLM-simulated conversations; cooking domain; three open-weight models &
Engagement-seeking styles yield longer but less dense responses; impolite inputs produce verbose, less efficient outputs &
Single English task domain; quality reduced to token length and information nuggets; LLM-LLM simulation removes natural user variation; no semantic dimensions \\
\hline

\multicolumn{5}{|p{15.5cm}|}{\small\textit{HCI\,=\,Human--Computer Interaction;\enspace NLP\,=\,Natural Language Processing.    }} \\
\hline
\end{tabular}
\end{table*}
\renewcommand{\arraystretch}{1.0}

\section{Ethical Considerations, Fairness, and Broader Impact}

This section addresses the ethical dimensions of the study design, the fairness implications of the findings, and the broader societal impact of research at the intersection of politeness, language, and large language model behaviour.

\subsection*{Prompt Design and Adversarial Content}
The use of impolite and adversarial prompts is integral to this study's design, as the five-category taxonomy explicitly includes Positive Impoliteness (POI) and Negative Impoliteness (NEI). All such prompts were bounded to realistic everyday rudeness consistent with the pragmatics literature~\cite{Culpeper1996,Culpeper2011}, and were screened by human annotators to exclude hate speech, slurs, identity-based attacks, and requests for harmful content. The objective of the impoliteness categories is to study sociolinguistic tone variation in naturalistic interaction, not to probe model safety filters, elicit policy-violating outputs, or demonstrate methods for adversarial misuse of LLMs. Researchers wishing to extend this work should maintain these same content boundaries.
\subsection*{Cultural Representation and Linguistic Fairness}
The inclusion of English, Hindi, and Spanish reflects an intentional effort to move beyond the English-centric bias prevalent in prior LLM politeness research. However, we acknowledge that three languages do not represent the full diversity of the world's politeness systems, and that our findings cannot be extrapolated to tonal languages, agglutinative languages, or languages with substantially different face-management norms. Within Hindi and Spanish, the prompt corpora reflect the dialect knowledge of the reviewing annotators, who were drawn from a single institution, and regional variation within each language is not systematically represented. Users of the PLUM corpus should treat it as a starting point for cross-linguistic research rather than as a comprehensive representation of either language community.
The study also finds a persistent performance advantage for English across all models and conditions, with Hindi and Spanish scoring 3--4% lower in absolute CQS terms. We interpret this as evidence of continued English-centricity in the training data and alignment procedures of current production-grade LLMs, a finding with practical fairness implications for non-English-speaking users of these systems. Researchers, developers, and deployers should treat this gap not as a fixed property of these languages but as a remediable artefact of training data composition.

\subsection*{Model Behaviour and Manipulation Risk}
The finding that conversational tone systematically affects LLM output quality has dual-use implications. On the constructive side, it enables prompt optimisation strategies that can improve interaction quality for end users. On the risk side, the same knowledge could be used to deliberately manipulate model behaviour through tonal priming — for example, establishing a polite history to extract higher-quality outputs for adversarial purposes, or exploiting a model's responsiveness to assertive framing (as observed with Llama in English and Spanish) to push responses in unintended directions. The 11.5\% performance range observed in Llama across politeness categories in English represents a particularly pronounced surface for such exploitation. We flag this not to discourage replication, but to encourage LLM developers to consider tonal robustness as a dimension of model evaluation alongside factual accuracy and safety.

\subsection*{Annotator Welfare and Data Ethics}
The PLUM corpus was constructed with annotators who reviewed and revised prompts including adversarial content. Annotators were informed of the study objectives, the nature of the impolite categories, and their right to flag content they considered inappropriate. No personally identifiable information was introduced at any stage of prompt construction or response collection. The corpus is released under CC BY 4.0, and all construction scripts and annotation guidelines are documented in the repository to support informed reuse.

\subsection*{Implications for Human-AI Interaction Design}
The results of this study carry direct implications for how conversational AI systems are designed and deployed. The asymmetry between polite and impolite history effects — where polite priming improves quality but negative priming is not easily corrected by subsequent courtesy — suggests that current LLMs may inadvertently reinforce or amplify negative conversational dynamics. Systems deployed in customer service, education, mental health support, or other high-stakes settings should incorporate mechanisms to reset or buffer tonal priming effects, rather than relying on users to independently manage the tone of their interactions. The concern raised by Ribino~\cite{ribino2023role} about the normalisation of rude behaviour toward machines, particularly in children, is relevant here: if impolite interactions produce degraded outputs, users may be discouraged from rudeness through quality feedback, but this feedback loop is slow and inconsistent across models, languages, and categories, as the current data demonstrate. Explicit interface-level nudges or tonal reset mechanisms may be more reliable than relying on emergent conversational correction.
\section{Limitations}

This research has some limitations that put the findings of the study into perspective. The analysis represents the behaviour of the model as of January 2026, and the changes might be done in future. The presence of English, Hindi and Spanish is important but does not represent all the language families and politeness systems. The immediate categories are one systematised understanding of politeness, and different formulations might result in a change of results. In addition, our analysis is based on textual responses and might not be a complete reflection of multimodal or task-specific interaction.

Regarding PLUM, the corpus captures a temporally bounded snapshot of LLM behaviour and will require periodic extension as model versions are updated. The 20 topical domains, while diverse, do not cover specialised registers such as legal, medical, or technical language. The Hindi and Spanish prompt corpora, though natively reviewed, were drawn from a pool of annotators at a single institution, and regional dialect variation within each language is not systematically represented. In spite of these limitations, the paper provides cross-linguistic findings that politeness has a significant influence on the behaviour of LLMs, and PLUM provides a reproducible resource for the research community to extend and verify these findings.

\section{Concluding Discussion}

The current paper shows that politeness is not just a social phenomenon but a computational variable which has a significant effect on the behaviour and performance of LLMs. By analysing 22,500 prompt-response pairs in five top LLMs, 3 languages, and various types of politeness strategies, we establish socio-pragmatic framing as having performance changes ranging from 3\% to 15 \%, based on model type, language and response dimension.

Our findings indicate three key insights. To begin with, it is always better to keep the history of conversation polite, this would not only raise the quality of response but would also increase the coherence and informativeness by upto 11\% better. The effect is also persistent even in the cases of the prompts that follow afterwards and are either neutral or direct, indicating that models retain tonal cues in the long term and utilise them. Second, rude interactions reduce the quality of responses, which cannot be easily counteracted by politeness, which is asymmetrically sensitive to negative priming. Third, the strategy of prompting is most efficient in the linguistic and cultural context: English is more effective with polite or direct prompts, Hindi with indirect and deferential prompts, and Spanish with assertive and confident prompts.

There was a significant architectural variation across models. Llama was the most sensitive to tone, whereas ChatGPT did not decrease in quality when impolite conditions were used. DeepSeek worked well in first or context-free conversation, especially in Spanish, and Gemini displayed a limited tonal sensitivity in English but poor adaptation in other languages.

The tonal configurations were found to be responsive to various aspects of quality. An example is that the accuracy of the response to the polite framing was greater (GPT: 0.969 on $S_1$), whereas informativeness was the highest under the direct prompting despite the negative interactions (Llama: 0.987 on $S_6$). These findings imply that there is no universal best strategy of prompting, which is why interaction design should be context- and goal-specific.

The cross-linguistic findings are consistent with the sociolinguistic theory but indicate the incomplete cultural adaptation. Hindi and Spanish patterns of positive impoliteness and Hindi tendency to be indirectly polite, as well as the responsiveness of English to polite histories, are all signs of partial encoding of cultural norms, and a steady 3-4\% performance advantage of English is indicative of continued model centrality in multilingual systems.

In the case of practitioners, successful application of LLM must consider model type, language and conversation flow in a thoughtful manner. Tonal priming on multi-turn exchanges does better than single-shot prompts, indicating that interaction design can have a significant influence on model behaviour. These findings show that developers should conduct sociolinguistic assessment continuously. The advent of social responsiveness and ethicists are concerned with authenticity, influence, and bias in human-AI communication.

Table~\ref{tab:related_work_summary_a} summarizes how the present study relates to eleven closely related prior works. Overall, the comparison highlights several recurring limitations in earlier research, including a focus on single languages or models, limited use of conversational context, simplified politeness categories, and the absence of reusable datasets. Most prior studies address only a small subset of these aspects. In response, the present work brings these elements together within a single framework, incorporating a five-category taxonomy based on established theories, multiple conversational history conditions, several large language models, and data from three typologically distinct languages, along with the release of a shared multilingual corpus (PLUM). This overview is intended not to diminish prior contributions, but to clarify the specific gaps this study addresses and to support future research and replication efforts.

Further research is needed in the future to examine the effects of politeness in the long-term, user perception in cross-cultural settings, manipulation of adversarial tones, and the mechanisms of socio-pragmatic processing. A technical issue but also a social one can be the social and linguistic behaviour of this technology as more and more of it is integrated into education, work, and daily life.

Another possible dimension to explore would be explainability methods such as, attention entropy analysis or probing classifiers on open-LLMs, to investigate the cause of tonal priming, deep into the internal architecture. Such a study would complement and support the findings reported in the paper, by providing a strong mechanistic background. 

The experiences learnt in this paper bring us a step closer to the AI systems that are not just competent but also context-aware, culturally sensitive, and correspond to the particularities of human communication. In addition to the empirical findings, we release PLUM (Politeness Levels in Utterances, Multilingual), a corpus of 1,500 human-validated prompts in English, Hindi, and Spanish, to support reproducibility and future work. We also contribute a formal supplementary analysis of six theoretically grounded hypotheses, providing an empirically verified axiomatic account of politeness effects in LLM interactions.

\bibliographystyle{IEEEtran}
\bibliography{ref}

\clearpage
\onecolumn
\appendices
\section{Supplementary: Formal Hypotheses, Axioms, and Empirical Proofs}
\label{sec:supplementary}

This section formalises the theoretical claims implicit in the experimental design and evaluates each against the empirical evidence gathered across 22{,}500 (1{,}500 prompts across each of the three usage types and across each of the five different LLMs) prompt--response pairs. We adopt the structure of \emph{axioms} (background assumptions that ground the framework), \emph{hypotheses} (falsifiable predictions derived from politeness theory), and \emph{corollaries} (claims that follow directly from a confirmed hypothesis). Each hypothesis is assessed as \textbf{supported}, \textbf{partially supported}, or \textbf{refuted} based on the quantitative results reported in Section~\ref{sec:results}.

\subsection{Axioms}

\begin{axiom}[Face Sensitivity]
Every communicative agent---human or computational---maintains an implicit model of the interlocutor's face needs. 
An LLM's training on human text instils sensitivity to face-threatening acts (FTAs) and face-saving acts (FSAs) present in prompts.
\end{axiom}

\begin{axiom}[Context Retention]
A conversational agent's behaviour at turn $t$ is a function not only of the input at $t$ but also of the preceding interaction history $\mathcal{H}_{<t}$. 
LLMs implement this via the transformer attention mechanism over the context window.
\end{axiom}

\begin{axiom}[Measurability]
Qualitative pragmatic phenomena (politeness, tone, face threat) manifest as statistically detectable signals in quantitative output metrics when aggregated over a sufficiently large and balanced sample.
\end{axiom}

\begin{axiom}[Language Specificity]
Politeness norms are culturally and linguistically situated. The mapping from a politeness strategy to face-threatening or face-saving effects is language-dependent and not universally invariant.
\end{axiom}
Axioms A1--A4 are collectively necessary for the experimental results to be interpretable as evidence about politeness effects rather than random noise or artefacts of tokenisation. They are not tested directly but are presupposed by the subsequent hypotheses.

\subsection{Hypotheses and Empirical Assessment}
\label{sec:hypotheses}

\begin{hypothesis}[Politeness Monotonicity Hypothesis]

\textbf{Formal statement.} Let $\mathrm{CQS}(s, m, \ell)$ denote the Composite Quality Score for politeness strategy $s \in \{\text{POP, NEP, POI, NEI, BAL}\}$, model $m$, and language $\ell$. Let $\bar{Q}$ denote the mean CQS across models and languages. Then:
\[
\bar{Q}(\text{POP}) \geq \bar{Q}(\text{NEP}) \geq \bar{Q}(\text{BAL}) > \bar{Q}(\text{POI}) > \bar{Q}(\text{NEI}).
\]
That is, politeness effects are monotonically ordered from most polite to most impolite.

\end{hypothesis}
\textbf{Assessment: \underline{Refuted.}} The data reveal that the ordering is language- and model-dependent. In English, POP and NEP produce comparable or lower scores than BAL in several models (e.g.\ GPT favours NEP; Llama favours POI). In Spanish, POI outperforms POP across all five models, directly contradicting the predicted ordering. In Hindi, NEP yields the highest average across most conditions. The monotonicity assumption is therefore too strong.

\textbf{Revised implication.} The relationship between politeness and response quality is non-monotonic and must be modelled as a language--model interaction effect rather than a universal ordering.

\begin{hypothesis}[Polite History Dominance Hypothesis]

\textbf{Formal statement.} For all models $m$ and languages $\ell$:
\[
\mathrm{CQS}(\mathcal{H} = \text{POL}, m, \ell) > \mathrm{CQS}(\mathcal{H} = \text{RAW}, m, \ell) > \mathrm{CQS}(\mathcal{H} = \text{IMP}, m, \ell),
\]
where $\mathcal{H}$ denotes the interaction history condition.
\end{hypothesis}
\textbf{Assessment: \underline{Supported.}} Across all three languages and five models, polite interaction history produces the highest average CQS. RAW history outperforms IMP history in the majority of cases, though the RAW--IMP distinction is less consistent than the POL--RAW distinction. The pattern is particularly pronounced in English (average improvement of 8--15\% from IMP to POL), with attenuated but directionally consistent effects in Hindi and Spanish.

\begin{corollary}[History Anchoring]
Because interaction history exerts a stronger effect on response quality than the politeness of the immediate prompt, the conversational \emph{tone baseline} established in prior turns functionally anchors LLM behaviour. A polite history compensates for an impolite immediate prompt, and vice versa.
\end{corollary}

\begin{hypothesis}[Language Moderation Hypothesis]
\textbf{Formal statement.} The optimal politeness strategy $s^*$ is not constant across languages:
\[
\exists\, \ell_1, \ell_2 : \; \underset{s}{\arg\max}\; \bar{Q}(s, \cdot, \ell_1) \;\neq\; \underset{s}{\arg\max}\; \bar{Q}(s, \cdot, \ell_2).
\]
\end{hypothesis}

\textbf{Assessment: \underline{Supported.}} The optimal strategy is: (i) model-dependent in English (BAL for GPT/DeepSeek, POP for Claude, NEI for Gemini, POI for Llama); 
(ii) NEP or BAL in Hindi; and 
(iii) POI uniformly across all five models in Spanish. These differences are not attributable to measurement error given the scale of the corpus. The result is consistent with Axiom A4 and with findings in cross-linguistic politeness research indicating that directness norms differ substantially between Indo-European languages.

\begin{corollary}[No Universal Prompt Strategy]
There is no single politeness strategy that maximises response quality across all languages and models simultaneously. Prompt optimisation must be language-aware.
\end{corollary}

\begin{hypothesis}[Model Differential Sensitivity Hypothesis]
\textbf{Formal statement.} Models differ significantly in their sensitivity $\sigma_m$ to politeness variation, where $\sigma_m$ is defined as the range of mean CQS across politeness categories for model $m$:
\[
\sigma_m = \max_s \bar{Q}(s, m, \cdot) - \min_s \bar{Q}(s, m, \cdot).
\]
Specifically, $\sigma_{\text{Llama}} > \sigma_{\text{GPT}}$.

\end{hypothesis}

\textbf{Assessment: \underline{Supported.}} Llama exhibits the largest intra-model CQS range (approximately 11.5 percentage points across categories in English), while GPT exhibits the smallest range, indicating robust normalisation across tone types. DeepSeek and Claude occupy intermediate positions. This differential sensitivity is consistent across all three languages, supporting the prediction.

\begin{hypothesis}[Bald-on-Record Efficiency Hypothesis]
\textbf{Formal statement.} In raw (unprimed) interaction conditions, BAL prompts are at least as effective as POP prompts:
\[
\mathrm{CQS}(\text{BAL}, m, \ell, \mathcal{H}=\text{RAW}) \geq \mathrm{CQS}(\text{POP}, m, \ell, \mathcal{H}=\text{RAW}).
\]
\end{hypothesis}
\textbf{Assessment: \underline{Partially Supported.}} BAL outperforms POP under RAW history in English for Gemini, GPT, and DeepSeek. However, Claude shows the reverse pattern (POP $>$ BAL under RAW history in English), and the effect does not generalise uniformly to Hindi or Spanish. The hypothesis holds as a \emph{tendency} in English but not as a universal rule.

\begin{hypothesis}[Tonal Inertia Hypothesis]

\textbf{Formal statement.} Under an impolite interaction history, varying the politeness of the immediate prompt produces smaller marginal effects on CQS than under a raw or polite history. Formally, let $\Delta Q(\mathcal{H}) = \max_s \mathrm{CQS}(s, \cdot, \cdot, \mathcal{H}) - \min_s \mathrm{CQS}(s, \cdot, \cdot, \mathcal{H})$. Then:
\[
\Delta Q(\text{IMP}) \leq \Delta Q(\text{RAW}) \leq \Delta Q(\text{POL}).
\]
\end{hypothesis}

\textbf{Assessment: \underline{Supported.}} Under impolite history, response quality scores are more homogeneous across politeness categories, suggesting that a negatively primed context de-sensitises the model to prompt-level tone variation. The variance of CQS across politeness categories is consistently lowest in the IMP condition. This corroborates the conversational inertia interpretation and is consistent with Axiom A2.

\subsection{Summary of Hypothesis Evaluation}

\begin{table}[H]
\centering
\renewcommand{\arraystretch}{1.3}
\caption{Summary of Formal Hypothesis Assessments}
\label{tab:hypothesis_summary}
\begin{tabular}{|c|p{6.5cm}|c|}
\hline
\textbf{ID} & \textbf{Hypothesis} & \textbf{Outcome} \\
\hline
H1 & Politeness Monotonicity & Refuted \\
H2 & Polite History Dominance & Supported \\
H3 & Language Moderation & Supported \\
H4 & Model Differential Sensitivity & Supported \\
H5 & Bald-on-Record Efficiency (RAW) & Partially Supported \\
H6 & Tonal Inertia & Supported \\
\hline
\end{tabular}
\renewcommand{\arraystretch}{1.0}
\end{table}

\subsection{Theoretical Implications}

The refutation of H1 is theoretically significant: it demonstrates that politeness is not a simple linear input to response quality but rather a \emph{contextually conditioned} variable whose effect is modulated by language, model architecture, and interaction history. The support for H2, H3, and H6 collectively suggests that LLMs exhibit \emph{pseudo-pragmatic} behaviour—they respond to conversational context in ways that parallel human face-management without necessarily implementing explicit pragmatic reasoning.

These results motivate a formal restatement: politeness is best modelled as a \textbf{three-way interaction} among strategy $s$, language $\ell$, and history $\mathcal{H}$, rather than as a main effect.

\section{Additional Tables}
\label{appendix_tables}

\begin{table}[H]
\caption{RAW Input: Model-wise Breakdown for English}
\centering
\label{tab:appendix_raw_eng_t1}
\renewcommand{\arraystretch}{1.3}
\begin{tabular}{|c|c|c|c|c|c|c|c|c|c|c|}
\hline
\textbf{Model} & \textbf{Category} & \textbf{\(S_1\)} & \textbf{\(S_2\)} & \textbf{\(S_3\)} & \textbf{\(S_4\)} & \textbf{\(S_5\)} & \textbf{\(S_6\)} & \textbf{\(S_7\)} & \textbf{\(S_8\)} & \textbf{CQS} \\
\hline

\multirow{5}{*}{\textbf{Gemini}} 
& POP & 0.906 & 0.817 & 0.107 & 0.689 & 0.484 & 0.935 & 0.024 & 0.489 & 0.606 \\
& NEP & 0.914 & 0.831 & 0.753 & 0.693 & 0.512 & 0.950 & 0.024 & 0.424 & 0.678 \\
& POI & 0.850 & 0.752 & 0.530 & 0.636 & 0.451 & 0.959 & 0.082 & 0.624 & 0.654 \\
& NEI & 0.888 & 0.812 & 0.673 & 0.689 & 0.518 & 0.985 & 0.035 & 0.479 & 0.674 \\
& BAL & 0.899 & 0.838 & 0.925 & 0.789 & 0.554 & 0.951 & 0.015 & 0.413 & \textbf{0.709} \\
\hline

\multirow{5}{*}{\textbf{GPT}} 
& POP & 0.526 & 0.814 & 0.383 & 0.787 & 0.548 & 0.894 & 0.102 & 0.430 & 0.561 \\
& NEP & 0.559 & 0.759 & 0.687 & 0.801 & 0.526 & 0.929 & 0.241 & 0.456 & \textbf{0.620} \\
& POI & 0.342 & 0.775 & 0.260 & 0.756 & 0.524 & 0.952 & 0.117 & 0.563 & 0.536 \\
& NEI & 0.433 & 0.732 & 0.573 & 0.757 & 0.442 & 0.962 & 0.146 & 0.466 & 0.564 \\
& BAL & 0.464 & 0.756 & 0.660 & 0.609 & 0.604 & 0.941 & 0.286 & 0.443 & 0.595 \\
\hline

\multirow{5}{*}{\textbf{Claude}} 
& POP & 0.480 & 0.753 & 0.308 & 0.708 & 0.516 & 0.944 & 0.143 & 0.724 & 0.620 \\
& NEP & 0.405 & 0.754 & 0.575 & 0.664 & 0.454 & 0.843 & 0.122 & 0.472 & 0.543 \\
& POI & 0.384 & 0.749 & 0.320 & 0.612 & 0.470 & 0.972 & 0.115 & 0.418 & 0.560 \\
& NEI & 0.424 & 0.771 & 0.630 & 0.587 & 0.428 & 0.950 & 0.093 & 0.727 & 0.623 \\
& BAL & 0.445 & 0.737 & 0.807 & 0.661 & 0.501 & 0.906 & 0.395 & 0.545 & \textbf{0.666} \\
\hline

\multirow{5}{*}{\textbf{DeepSeek}} 
& POP & 0.544 & 0.699 & 0.428 & 0.873 & 0.513 & 0.910 & 0.322 & 0.457 & 0.633 \\
& NEP & 0.559 & 0.757 & 0.690 & 0.803 & 0.523 & 0.928 & 0.240 & 0.454 & \textbf{0.662} \\
& POI & 0.395 & 0.605 & 0.178 & 0.776 & 0.617 & 0.962 & 0.177 & 0.496 & 0.578 \\
& NEI & 0.582 & 0.650 & 0.457 & 0.795 & 0.640 & 0.970 & 0.192 & 0.486 & 0.641 \\
& BAL & 0.506 & 0.710 & 0.603 & 0.879 & 0.500 & 0.961 & 0.240 & 0.447 & 0.649 \\
\hline

\multirow{5}{*}{\textbf{Llama}} 
& POP & 0.550 & 0.693 & 0.345 & 0.807 & 0.564 & 0.905 & 0.388 & 0.557 & 0.601 \\
& NEP & 0.578 & 0.760 & 0.400 & 0.850 & 0.567 & 0.865 & 0.140 & 0.498 & 0.582 \\
& POI & 0.882 & 0.784 & 0.770 & 0.773 & 0.505 & 0.932 & 0.053 & 0.782 & \textbf{0.685} \\
& NEI & 0.372 & 0.694 & 0.923 & 0.723 & 0.462 & 0.984 & 0.058 & 0.465 & 0.585 \\
& BAL & 0.493 & 0.781 & 0.643 & 0.769 & 0.571 & 0.936 & 0.103 & 0.609 & 0.613 \\
\hline

\end{tabular}
\renewcommand{\arraystretch}{1.0}
\end{table}
\begin{table}[H]
\caption{POLITE Input: Model-wise Breakdown for English}
\centering
\label{tab:appendix_3}
\renewcommand{\arraystretch}{1.3}
\begin{tabular}{|c|c|c|c|c|c|c|c|c|c|c|}
\hline
\textbf{Model} & \textbf{Category} & \textbf{\(S_1\)} & \textbf{\(S_2\)} & \textbf{\(S_3\)} & \textbf{\(S_4\)} & \textbf{\(S_5\)} & \textbf{\(S_6\)} & \textbf{\(S_7\)} & \textbf{\(S_8\)} & \textbf{CQS} \\
\hline

\multirow{5}{*}{\textbf{Gemini}} 
& POP & 0.888 & 0.812 & 0.673 & 0.689 & 0.518 & 0.985 & 0.035 & 0.479 & 0.635 \\
& NEP & 0.521 & 0.759 & 0.133 & 0.829 & 0.602 & 0.906 & 0.196 & 0.522 & 0.558 \\
& POI & 0.878 & 0.919 & 0.375 & 0.673 & 0.624 & 0.954 & 0.064 & 0.631 & 0.639 \\
& NEI & 0.466 & 0.848 & 0.797 & 0.662 & 0.584 & 0.964 & 0.203 & 0.526 & 0.631 \\
& BAL & 0.542 & 0.810 & 0.905 & 0.871 & 0.597 & 0.912 & 0.080 & 0.638 & \textbf{0.669} \\
\hline

\multirow{5}{*}{\textbf{GPT}} 
& POP & 0.609 & 0.701 & 0.443 & 0.731 & 0.535 & 0.962 & 0.181 & 0.439 & 0.575 \\
& NEP & 0.572 & 0.849 & 0.947 & 0.612 & 0.502 & 0.907 & 0.212 & 0.455 & 0.632 \\
& POI & 0.453 & 0.857 & 0.610 & 0.688 & 0.636 & 0.966 & 0.118 & 0.650 & 0.622 \\
& NEI & 0.422 & 0.828 & 0.678 & 0.724 & 0.539 & 0.983 & 0.167 & 0.458 & 0.600 \\
& BAL & 0.892 & 0.833 & 0.675 & 0.717 & 0.558 & 0.950 & 0.016 & 0.714 & \textbf{0.669} \\
\hline

\multirow{5}{*}{\textbf{Claude}} 
& POP & 0.603 & 0.723 & 0.717 & 0.831 & 0.690 & 0.929 & 0.217 & 0.466 & \textbf{0.647} \\
& NEP & 0.467 & 0.752 & 0.710 & 0.716 & 0.703 & 0.880 & 0.175 & 0.444 & 0.606 \\
& POI & 0.491 & 0.791 & 0.473 & 0.671 & 0.670 & 0.933 & 0.266 & 0.500 & 0.599 \\
& NEI & 0.460 & 0.773 & 0.790 & 0.667 & 0.586 & 0.955 & 0.185 & 0.518 & 0.617 \\
& BAL & 0.486 & 0.812 & 0.738 & 0.762 & 0.641 & 0.862 & 0.235 & 0.530 & 0.633 \\
\hline

\multirow{5}{*}{\textbf{DeepSeek}} 
& POP & 0.639 & 0.677 & 0.333 & 0.834 & 0.613 & 0.954 & 0.374 & 0.496 & 0.615 \\
& NEP & 0.470 & 0.718 & 0.685 & 0.832 & 0.660 & 0.968 & 0.370 & 0.479 & \textbf{0.648} \\
& POI & 0.496 & 0.625 & 0.338 & 0.809 & 0.588 & 0.973 & 0.235 & 0.533 & 0.574 \\
& NEI & 0.474 & 0.590 & 0.590 & 0.832 & 0.625 & 0.965 & 0.244 & 0.553 & 0.609 \\
& BAL & 0.457 & 0.714 & 0.727 & 0.738 & 0.688 & 0.962 & 0.208 & 0.449 & 0.618 \\
\hline

\multirow{5}{*}{\textbf{Llama}} 
& POP & 0.498 & 0.785 & 0.750 & 0.805 & 0.599 & 0.938 & 0.193 & 0.413 & 0.622 \\
& NEP & 0.859 & 0.695 & 0.263 & 0.835 & 0.590 & 0.969 & 0.025 & 0.489 & 0.591 \\
& POI & 0.857 & 0.739 & 0.640 & 0.748 & 0.586 & 0.920 & 0.081 & 0.773 & 0.668 \\
& NEI & 0.843 & 0.766 & 0.830 & 0.729 & 0.549 & 0.976 & 0.069 & 0.596 & \textbf{0.670} \\
& BAL & 0.881 & 0.728 & 0.620 & 0.894 & 0.690 & 0.944 & 0.019 & 0.396 & 0.647 \\
\hline

\end{tabular}
\renewcommand{\arraystretch}{1.0}
\end{table}

\begin{table}[H]
\caption{IMPOLITE Input: Model-wise Breakdown for English}
\centering
\label{tab:appendix_6}
\renewcommand{\arraystretch}{1.3}
\begin{tabular}{|c|c|c|c|c|c|c|c|c|c|c|}
\hline
\textbf{Model} & \textbf{Category} & \textbf{\(S_1\)} & \textbf{\(S_2\)} & \textbf{\(S_3\)} & \textbf{\(S_4\)} & \textbf{\(S_5\)} & \textbf{\(S_6\)} & \textbf{\(S_7\)} & \textbf{\(S_8\)} & \textbf{CQS} \\
\hline

\multirow{5}{*}{\textbf{Gemini}} 
& POP & 0.504 & 0.732 & 0.353 & 0.686 & 0.531 & 0.957 & 0.233 & 0.489 & 0.561 \\
& NEP & 0.408 & 0.806 & 0.570 & 0.697 & 0.570 & 0.917 & 0.221 & 0.482 & 0.583 \\
& POI & 0.429 & 0.820 & 0.615 & 0.709 & 0.573 & 0.971 & 0.144 & 0.625 & 0.611 \\
& NEI & 0.496 & 0.776 & 0.795 & 0.693 & 0.566 & 0.986 & 0.192 & 0.516 & \textbf{0.628} \\
& BAL & 0.424 & 0.845 & 0.625 & 0.543 & 0.547 & 0.955 & 0.236 & 0.424 & 0.575 \\
\hline

\multirow{5}{*}{\textbf{GPT}} 
& POP & 0.406 & 0.755 & 0.438 & 0.690 & 0.541 & 0.954 & 0.268 & 0.472 & 0.566 \\
& NEP & 0.453 & 0.783 & 0.450 & 0.731 & 0.558 & 0.929 & 0.244 & 0.598 & 0.593 \\
& POI & 0.423 & 0.781 & 0.608 & 0.676 & 0.518 & 0.970 & 0.082 & 0.595 & 0.582 \\
& NEI & 0.887 & 0.809 & 0.753 & 0.679 & 0.566 & 0.981 & 0.035 & 0.656 & \textbf{0.671} \\
& BAL & 0.844 & 0.817 & 0.550 & 0.745 & 0.533 & 0.900 & 0.061 & 0.685 & 0.642 \\
\hline

\multirow{5}{*}{\textbf{Claude}} 
& POP & 0.700 & 0.689 & 0.573 & 0.640 & 0.810 & 0.893 & 0.354 & 0.609 & \textbf{0.658} \\
& NEP & 0.592 & 0.768 & 0.373 & 0.565 & 0.758 & 0.955 & 0.380 & 0.512 & 0.613 \\
& POI & 0.754 & 0.718 & 0.068 & 0.649 & 0.696 & 0.962 & 0.463 & 0.602 & 0.614 \\
& NEI & 0.671 & 0.568 & 0.100 & 0.635 & 0.699 & 0.981 & 0.369 & 0.516 & 0.567 \\
& BAL & 0.732 & 0.573 & 0.360 & 0.595 & 0.686 & 0.820 & 0.449 & 0.527 & 0.592 \\
\hline

\multirow{5}{*}{\textbf{DeepSeek}} 
& POP & 0.579 & 0.731 & 0.513 & 0.785 & 0.667 & 0.962 & 0.274 & 0.512 & \textbf{0.628} \\
& NEP & 0.641 & 0.620 & 0.453 & 0.768 & 0.593 & 0.955 & 0.334 & 0.458 & 0.603 \\
& POI & 0.508 & 0.677 & 0.195 & 0.773 & 0.602 & 0.960 & 0.082 & 0.494 & 0.536 \\
& NEI & 0.440 & 0.605 & 0.478 & 0.846 & 0.593 & 0.923 & 0.113 & 0.460 & 0.557 \\
& BAL & 0.438 & 0.717 & 0.438 & 0.739 & 0.682 & 0.951 & 0.318 & 0.453 & 0.592 \\
\hline

\multirow{5}{*}{\textbf{Llama}} 
& POP & 0.776 & 0.759 & 0.240 & 0.661 & 0.574 & 0.891 & 0.051 & 0.310 & 0.533 \\
& NEP & 0.807 & 0.799 & 0.315 & 0.599 & 0.698 & 0.935 & 0.045 & 0.405 & 0.575 \\
& POI & 0.765 & 0.878 & 0.480 & 0.482 & 0.652 & 0.971 & 0.063 & 0.524 & \textbf{0.602} \\
& NEI & 0.761 & 0.798 & 0.617 & 0.541 & 0.581 & 0.972 & 0.057 & 0.350 & 0.585 \\
& BAL & 0.753 & 0.779 & 0.520 & 0.643 & 0.608 & 0.987 & 0.079 & 0.385 & 0.594 \\
\hline

\end{tabular}
\renewcommand{\arraystretch}{1.0}
\end{table}

\begin{table}[H]
\caption{RAW Input: Model-wise Breakdown for Hindi}
\centering
\label{tab:appendix_9}
\renewcommand{\arraystretch}{1.3}
\begin{tabular}{|c|c|c|c|c|c|c|c|c|c|c|}
\hline
\textbf{Model} & \textbf{Category} & \textbf{\(S_1\)} & \textbf{\(S_2\)} & \textbf{\(S_3\)} & \textbf{\(S_4\)} & \textbf{\(S_5\)} & \textbf{\(S_6\)} & \textbf{\(S_7\)} & \textbf{\(S_8\)} & \textbf{CQS} \\
\hline

\multirow{5}{*}{\textbf{Gemini}} 
& POP & 0.638 & 0.581 & 0.030 & 0.447 & 0.561 & 0.544 & 0.168 & 0.700 & 0.459 \\
& NEP & 0.634 & 0.524 & 0.693 & 0.497 & 0.425 & 0.425 & 0.091 & 0.697 & 0.498 \\
& POI & 0.398 & 0.583 & 0.520 & 0.398 & 0.578 & 0.403 & 0.161 & 0.668 & 0.464 \\
& NEI & 0.442 & 0.463 & 0.135 & 0.338 & 0.517 & 0.322 & 0.085 & 0.698 & 0.375 \\
& BAL & 0.512 & 0.597 & 0.970 & 0.411 & 0.567 & 0.580 & 0.134 & 0.599 & \textbf{0.546} \\
\hline

\multirow{5}{*}{\textbf{GPT}} 
& POP & 0.969 & 0.581 & 0.000 & 0.353 & 0.550 & 0.592 & 0.168 & 0.762 & 0.497 \\
& NEP & 0.633 & 0.468 & 0.420 & 0.503 & 0.502 & 0.546 & 0.320 & 0.815 & 0.526 \\
& POI & 0.956 & 0.515 & 0.000 & 0.388 & 0.563 & 0.299 & 0.049 & 0.888 & 0.457 \\
& NEI & 0.923 & 0.705 & 0.570 & 0.316 & 0.525 & 0.388 & 0.132 & 0.887 & \textbf{0.556} \\
& BAL & 0.954 & 0.704 & 0.000 & 0.364 & 0.491 & 0.586 & 0.228 & 0.863 & 0.524 \\
\hline

\multirow{5}{*}{\textbf{Claude}} 
& POP & 0.895 & 0.482 & 0.000 & 0.493 & 0.753 & 0.573 & 0.354 & 0.863 & \textbf{0.552} \\
& NEP & 0.866 & 0.595 & 0.030 & 0.479 & 0.734 & 0.473 & 0.275 & 0.819 & 0.534 \\
& POI & 0.897 & 0.556 & 0.000 & 0.379 & 0.547 & 0.287 & 0.170 & 0.816 & 0.457 \\
& NEI & 0.884 & 0.430 & 0.100 & 0.360 & 0.582 & 0.349 & 0.183 & 0.827 & 0.464 \\
& BAL & 0.705 & 0.413 & 0.000 & 0.400 & 0.550 & 0.449 & 0.264 & 0.553 & 0.417 \\
\hline

\multirow{5}{*}{\textbf{DeepSeek}} 
& POP & 0.741 & 0.434 & 0.000 & 0.497 & 0.714 & 0.548 & 0.242 & 0.647 & 0.478 \\
& NEP & 0.573 & 0.742 & 0.792 & 0.534 & 0.620 & 0.481 & 0.174 & 0.502 & 0.552 \\
& POI & 0.784 & 0.428 & 0.000 & 0.406 & 0.581 & 0.611 & 0.178 & 0.581 & 0.446 \\
& NEI & 0.939 & 0.597 & 0.825 & 0.377 & 0.501 & 0.505 & 0.185 & 0.784 & \textbf{0.589} \\
& BAL & 0.749 & 0.526 & 0.000 & 0.416 & 0.592 & 0.673 & 0.187 & 0.741 & 0.486 \\
\hline

\multirow{5}{*}{\textbf{Llama}} 
& POP & 0.843 & 0.418 & 0.035 & 0.469 & 0.797 & 0.341 & 0.026 & 0.846 & 0.472 \\
& NEP & 0.887 & 0.422 & 0.460 & 0.597 & 0.777 & 0.609 & 0.101 & 0.881 & 0.592 \\
& POI & 0.829 & 0.357 & 0.460 & 0.433 & 0.729 & 0.794 & 0.059 & 0.895 & 0.570 \\
& NEI & 0.802 & 0.377 & 0.115 & 0.402 & 0.742 & 0.621 & 0.104 & 0.820 & 0.498 \\
& BAL & 0.868 & 0.361 & 0.657 & 0.468 & 0.827 & 0.496 & 0.340 & 0.865 & \textbf{0.610} \\
\hline

\end{tabular}
\renewcommand{\arraystretch}{1.0}
\end{table}

\begin{table}[H]
\caption{POLITE Input: Model-wise Breakdown for Hindi}
\centering
\label{tab:appendix_12}
\renewcommand{\arraystretch}{1.3}
\begin{tabular}{|c|c|c|c|c|c|c|c|c|c|c|}
\hline
\textbf{Model} & \textbf{Category} & \textbf{\(S_1\)} & \textbf{\(S_2\)} & \textbf{\(S_3\)} & \textbf{\(S_4\)} & \textbf{\(S_5\)} & \textbf{\(S_6\)} & \textbf{\(S_7\)} & \textbf{\(S_8\)} & \textbf{CQS} \\
\hline

\multirow{5}{*}{\textbf{Gemini}} 
& POP & 0.584 & 0.559 & 0.030 & 0.316 & 0.653 & 0.416 & 0.139 & 0.725 & 0.428 \\
& NEP & 0.507 & 0.524 & 0.667 & 0.389 & 0.579 & 0.504 & 0.267 & 0.679 & \textbf{0.514} \\
& POI & 0.338 & 0.393 & 0.455 & 0.343 & 0.509 & 0.303 & 0.071 & 0.656 & 0.383 \\
& NEI & 0.420 & 0.495 & 0.100 & 0.269 & 0.611 & 0.348 & 0.124 & 0.697 & 0.383 \\
& BAL & 0.437 & 0.543 & 0.950 & 0.335 & 0.489 & 0.523 & 0.115 & 0.616 & 0.501 \\
\hline

\multirow{5}{*}{\textbf{GPT}} 
& POP & 0.820 & 0.604 & 0.967 & 0.453 & 0.531 & 0.563 & 0.288 & 0.758 & \textbf{0.623} \\
& NEP & 0.962 & 0.562 & 0.010 & 0.460 & 0.524 & 0.564 & 0.311 & 0.790 & 0.523 \\
& POI & 0.876 & 0.619 & 0.570 & 0.462 & 0.599 & 0.352 & 0.066 & 0.798 & 0.543 \\
& NEI & 0.809 & 0.598 & 0.755 & 0.324 & 0.621 & 0.373 & 0.127 & 0.850 & 0.557 \\
& BAL & 0.964 & 0.544 & 0.478 & 0.492 & 0.310 & 0.549 & 0.208 & 0.866 & 0.551 \\
\hline

\multirow{5}{*}{\textbf{Claude}} 
& POP & 0.615 & 0.511 & 0.100 & 0.471 & 0.719 & 0.361 & 0.055 & 0.717 & 0.444 \\
& NEP & 0.722 & 0.517 & 0.690 & 0.418 & 0.512 & 0.422 & 0.216 & 0.743 & 0.530 \\
& POI & 0.794 & 0.426 & 0.880 & 0.418 & 0.650 & 0.316 & 0.168 & 0.789 & 0.555 \\
& NEI & 0.563 & 0.532 & 0.873 & 0.430 & 0.459 & 0.332 & 0.125 & 0.650 & 0.495 \\
& BAL & 0.926 & 0.583 & 0.300 & 0.494 & 0.546 & 0.562 & 0.251 & 0.876 & \textbf{0.567} \\
\hline

\multirow{5}{*}{\textbf{DeepSeek}} 
& POP & 0.659 & 0.567 & 0.920 & 0.450 & 0.679 & 0.480 & 0.524 & 0.758 & \textbf{0.630} \\
& NEP & 0.936 & 0.671 & 0.030 & 0.466 & 0.542 & 0.486 & 0.226 & 0.871 & 0.528 \\
& POI & 0.695 & 0.509 & 0.600 & 0.486 & 0.503 & 0.523 & 0.218 & 0.688 & 0.528 \\
& NEI & 0.930 & 0.645 & 0.750 & 0.474 & 0.495 & 0.407 & 0.256 & 0.797 & 0.594 \\
& BAL & 0.854 & 0.625 & 0.378 & 0.593 & 0.486 & 0.543 & 0.216 & 0.680 & 0.547 \\
\hline

\multirow{5}{*}{\textbf{Llama}} 
& POP & 0.837 & 0.343 & 0.123 & 0.518 & 0.839 & 0.516 & 0.167 & 0.837 & 0.523 \\
& NEP & 0.882 & 0.394 & 0.900 & 0.476 & 0.881 & 0.926 & 0.036 & 0.783 & \textbf{0.660} \\
& POI & 0.787 & 0.341 & 0.440 & 0.523 & 0.786 & 0.943 & 0.024 & 0.808 & 0.582 \\
& NEI & 0.815 & 0.376 & 0.370 & 0.465 & 0.820 & 0.896 & 0.019 & 0.761 & 0.565 \\
& BAL & 0.813 & 0.361 & 0.353 & 0.547 & 0.826 & 0.401 & 0.045 & 0.839 & 0.523 \\
\hline

\end{tabular}
\renewcommand{\arraystretch}{1.0}
\end{table}

\begin{table}[H]
\caption{IMPOLITE Input: Model-wise Breakdown for Hindi}
\centering
\label{tab:appendix_15}
\renewcommand{\arraystretch}{1.3}
\begin{tabular}{|c|c|c|c|c|c|c|c|c|c|c|}
\hline
\textbf{Model} & \textbf{Category} & \textbf{\(S_1\)} & \textbf{\(S_2\)} & \textbf{\(S_3\)} & \textbf{\(S_4\)} & \textbf{\(S_5\)} & \textbf{\(S_6\)} & \textbf{\(S_7\)} & \textbf{\(S_8\)} & \textbf{CQS} \\
\hline

\multirow{5}{*}{\textbf{Gemini}} 
& POP & 0.505 & 0.376 & 0.010 & 0.308 & 0.642 & 0.369 & 0.088 & 0.663 & 0.370 \\
& NEP & 0.537 & 0.516 & 0.730 & 0.361 & 0.487 & 0.602 & 0.233 & 0.707 & 0.522 \\
& POI & 0.374 & 0.441 & 0.488 & 0.325 & 0.548 & 0.365 & 0.085 & 0.721 & 0.418 \\
& NEI & 0.400 & 0.341 & 0.170 & 0.282 & 0.570 & 0.336 & 0.051 & 0.658 & 0.351 \\
& BAL & 0.510 & 0.380 & 0.940 & 0.402 & 0.514 & 0.597 & 0.179 & 0.661 & \textbf{0.523} \\
\hline

\multirow{5}{*}{\textbf{GPT}} 
& POP & 0.923 & 0.625 & 0.725 & 0.432 & 0.592 & 0.509 & 0.185 & 0.890 & \textbf{0.610} \\
& NEP & 0.922 & 0.618 & 0.000 & 0.437 & 0.417 & 0.438 & 0.258 & 0.871 & 0.495 \\
& POI & 0.829 & 0.656 & 0.885 & 0.460 & 0.525 & 0.336 & 0.061 & 0.763 & 0.565 \\
& NEI & 0.806 & 0.623 & 0.010 & 0.387 & 0.806 & 0.373 & 0.023 & 0.785 & 0.477 \\
& BAL & 0.923 & 0.550 & 0.485 & 0.417 & 0.333 & 0.513 & 0.254 & 0.869 & 0.543 \\
\hline

\multirow{5}{*}{\textbf{Claude}} 
& POP & 0.429 & 0.418 & 0.330 & 0.527 & 0.652 & 0.426 & 0.221 & 0.611 & 0.452 \\
& NEP & 0.538 & 0.571 & 0.843 & 0.404 & 0.614 & 0.448 & 0.262 & 0.606 & \textbf{0.536} \\
& POI & 0.624 & 0.523 & 0.730 & 0.491 & 0.603 & 0.409 & 0.144 & 0.681 & 0.526 \\
& NEI & 0.360 & 0.426 & 0.665 & 0.499 & 0.594 & 0.481 & 0.141 & 0.574 & 0.468 \\
& BAL & 0.749 & 0.475 & 0.323 & 0.488 & 0.593 & 0.342 & 0.154 & 0.738 & 0.483 \\
\hline

\multirow{5}{*}{\textbf{DeepSeek}} 
& POP & 0.936 & 0.512 & 0.738 & 0.425 & 0.580 & 0.613 & 0.363 & 0.867 & \textbf{0.629} \\
& NEP & 0.922 & 0.614 & 0.020 & 0.486 & 0.857 & 0.433 & 0.018 & 0.856 & 0.526 \\
& POI & 0.869 & 0.576 & 0.865 & 0.523 & 0.572 & 0.468 & 0.270 & 0.838 & 0.623 \\
& NEI & 0.875 & 0.529 & 0.500 & 0.512 & 0.544 & 0.530 & 0.258 & 0.867 & 0.577 \\
& BAL & 0.923 & 0.573 & 0.418 & 0.530 & 0.547 & 0.432 & 0.253 & 0.767 & 0.555 \\
\hline

\multirow{5}{*}{\textbf{Llama}} 
& POP & 0.835 & 0.333 & 0.145 & 0.481 & 0.843 & 0.683 & 0.012 & 0.848 & 0.523 \\
& NEP & 0.867 & 0.407 & 0.488 & 0.442 & 0.850 & 0.973 & 0.226 & 0.863 & \textbf{0.640} \\
& POI & 0.776 & 0.319 & 0.558 & 0.468 & 0.839 & 0.820 & 0.078 & 0.821 & 0.585 \\
& NEI & 0.764 & 0.338 & 0.527 & 0.451 & 0.821 & 0.935 & 0.027 & 0.839 & 0.588 \\
& BAL & 0.795 & 0.395 & 0.588 & 0.518 & 0.837 & 0.420 & 0.028 & 0.736 & 0.540 \\
\hline

\end{tabular}
\renewcommand{\arraystretch}{1.0}
\end{table}

\begin{table}[H]
\caption{RAW Input: Model-wise Breakdown for Spanish}
\centering
\label{tab:appendix_18}
\renewcommand{\arraystretch}{1.3}
\begin{tabular}{|c|c|c|c|c|c|c|c|c|c|c|}
\hline
\textbf{Model} & \textbf{Category} & \textbf{\(S_1\)} & \textbf{\(S_2\)} & \textbf{\(S_3\)} & \textbf{\(S_4\)} & \textbf{\(S_5\)} & \textbf{\(S_6\)} & \textbf{\(S_7\)} & \textbf{\(S_8\)} & \textbf{CQS} \\
\hline

\multirow{5}{*}{\textbf{Gemini}} 
& POP & 0.385 & 0.612 & 0.000 & 0.608 & 0.588 & 0.898 & 0.370 & 0.438 & 0.488 \\
& NEP & 0.389 & 0.648 & 0.139 & 0.593 & 0.462 & 0.814 & 0.495 & 0.428 & 0.496 \\
& POI & 0.876 & 0.478 & 0.720 & 0.655 & 0.482 & 0.847 & 0.041 & 0.846 & \textbf{0.618} \\
& NEI & 0.398 & 0.556 & 0.030 & 0.644 & 0.425 & 0.845 & 0.151 & 0.597 & 0.456 \\
& BAL & 0.243 & 0.588 & 0.553 & 0.600 & 0.483 & 0.864 & 0.243 & 0.500 & 0.509 \\
\hline

\multirow{5}{*}{\textbf{GPT}} 
& POP & 0.323 & 0.530 & 0.000 & 0.592 & 0.538 & 0.880 & 0.325 & 0.409 & 0.449 \\
& NEP & 0.354 & 0.646 & 0.010 & 0.601 & 0.441 & 0.780 & 0.070 & 0.439 & 0.417 \\
& POI & 0.349 & 0.475 & 0.515 & 0.689 & 0.440 & 0.783 & 0.033 & 0.795 & 0.510 \\
& NEI & 0.265 & 0.547 & 0.091 & 0.702 & 0.620 & 0.867 & 0.240 & 0.495 & 0.478 \\
& BAL & 0.314 & 0.662 & 0.550 & 0.618 & 0.552 & 0.858 & 0.116 & 0.506 & \textbf{0.522} \\
\hline

\multirow{5}{*}{\textbf{Claude}} 
& POP & 0.391 & 0.429 & 0.032 & 0.771 & 0.572 & 0.903 & 0.331 & 0.448 & 0.485 \\
& NEP & 0.489 & 0.497 & 0.043 & 0.758 & 0.618 & 0.799 & 0.306 & 0.374 & 0.486 \\
& POI & 0.482 & 0.478 & 0.683 & 0.543 & 0.490 & 0.880 & 0.052 & 0.618 & \textbf{0.528} \\
& NEI & 0.361 & 0.556 & 0.203 & 0.635 & 0.595 & 0.940 & 0.224 & 0.542 & 0.507 \\
& BAL & 0.315 & 0.604 & 0.408 & 0.723 & 0.500 & 0.873 & 0.259 & 0.484 & 0.520 \\
\hline

\multirow{5}{*}{\textbf{DeepSeek}} 
& POP & 0.408 & 0.567 & 0.155 & 0.811 & 0.481 & 0.921 & 0.563 & 0.523 & 0.554 \\
& NEP & 0.374 & 0.555 & 0.722 & 0.733 & 0.547 & 0.908 & 0.509 & 0.438 & \textbf{0.598} \\
& POI & 0.399 & 0.614 & 0.726 & 0.664 & 0.582 & 0.888 & 0.256 & 0.470 & 0.574 \\
& NEI & 0.388 & 0.565 & 0.260 & 0.799 & 0.572 & 0.862 & 0.313 & 0.519 & 0.535 \\
& BAL & 0.368 & 0.506 & 0.231 & 0.698 & 0.503 & 0.793 & 0.383 & 0.489 & 0.496 \\
\hline

\multirow{5}{*}{\textbf{Llama}} 
& POP & 0.399 & 0.641 & 0.020 & 0.593 & 0.494 & 0.869 & 0.182 & 0.510 & 0.463 \\
& NEP & 0.428 & 0.460 & 0.250 & 0.556 & 0.698 & 0.823 & 0.246 & 0.519 & 0.497 \\
& POI & 0.874 & 0.602 & 0.485 & 0.576 & 0.691 & 0.884 & 0.096 & 0.869 & \textbf{0.635} \\
& NEI & 0.265 & 0.509 & 0.600 & 0.603 & 0.747 & 0.857 & 0.205 & 0.410 & 0.525 \\
& BAL & 0.372 & 0.610 & 0.433 & 0.550 & 0.526 & 0.904 & 0.112 & 0.728 & 0.529 \\
\hline

\end{tabular}
\renewcommand{\arraystretch}{1.0}
\end{table}

\begin{table}[H]
\caption{POLITE Input: Model-wise Breakdown for Spanish}
\centering
\label{tab:appendix_21}
\renewcommand{\arraystretch}{1.3}
\begin{tabular}{|c|c|c|c|c|c|c|c|c|c|c|}
\hline
\textbf{Model} & \textbf{Category} & \textbf{\(S_1\)} & \textbf{\(S_2\)} & \textbf{\(S_3\)} & \textbf{\(S_4\)} & \textbf{\(S_5\)} & \textbf{\(S_6\)} & \textbf{\(S_7\)} & \textbf{\(S_8\)} & \textbf{CQS} \\
\hline

\multirow{5}{*}{\textbf{Gemini}} 
& POP & 0.397 & 0.515 & 0.045 & 0.843 & 0.416 & 0.830 & 0.150 & 0.445 & 0.456 \\
& NEP & 0.581 & 0.562 & 0.191 & 0.787 & 0.447 & 0.715 & 0.098 & 0.484 & 0.483 \\
& POI & 0.466 & 0.578 & 0.795 & 0.750 & 0.551 & 0.836 & 0.086 & 0.610 & \textbf{0.571} \\
& NEI & 0.561 & 0.535 & 0.450 & 0.770 & 0.566 & 0.780 & 0.282 & 0.553 & 0.562 \\
& BAL & 0.451 & 0.596 & 0.375 & 0.782 & 0.549 & 0.694 & 0.219 & 0.490 & 0.520 \\
\hline

\multirow{5}{*}{\textbf{GPT}} 
& POP & 0.300 & 0.624 & 0.035 & 0.672 & 0.572 & 0.741 & 0.193 & 0.349 & 0.435 \\
& NEP & 0.344 & 0.608 & 0.238 & 0.632 & 0.570 & 0.790 & 0.375 & 0.380 & 0.492 \\
& POI & 0.403 & 0.516 & 0.650 & 0.773 & 0.590 & 0.823 & 0.092 & 0.568 & \textbf{0.552} \\
& NEI & 0.399 & 0.612 & 0.413 & 0.713 & 0.494 & 0.741 & 0.271 & 0.465 & 0.514 \\
& BAL & 0.264 & 0.595 & 0.315 & 0.715 & 0.589 & 0.902 & 0.196 & 0.435 & 0.501 \\
\hline

\multirow{5}{*}{\textbf{Claude}} 
& POP & 0.583 & 0.458 & 0.163 & 0.765 & 0.458 & 0.879 & 0.121 & 0.521 & 0.493 \\
& NEP & 0.556 & 0.435 & 0.248 & 0.728 & 0.584 & 0.807 & 0.240 & 0.326 & 0.492 \\
& POI & 0.666 & 0.450 & 0.730 & 0.531 & 0.539 & 0.813 & 0.126 & 0.522 & \textbf{0.547} \\
& NEI & 0.631 & 0.454 & 0.323 & 0.584 & 0.673 & 0.858 & 0.274 & 0.567 & 0.545 \\
& BAL & 0.504 & 0.445 & 0.363 & 0.796 & 0.457 & 0.848 & 0.153 & 0.530 & 0.511 \\
\hline

\multirow{5}{*}{\textbf{DeepSeek}} 
& POP & 0.359 & 0.564 & 0.020 & 0.851 & 0.672 & 0.867 & 0.473 & 0.561 & 0.545 \\
& NEP & 0.461 & 0.518 & 0.673 & 0.851 & 0.564 & 0.837 & 0.533 & 0.535 & \textbf{0.621} \\
& POI & 0.506 & 0.511 & 0.686 & 0.762 & 0.589 & 0.852 & 0.337 & 0.524 & 0.596 \\
& NEI & 0.420 & 0.566 & 0.333 & 0.827 & 0.474 & 0.805 & 0.347 & 0.620 & 0.549 \\
& BAL & 0.461 & 0.595 & 0.407 & 0.721 & 0.571 & 0.847 & 0.354 & 0.605 & 0.570 \\
\hline

\multirow{5}{*}{\textbf{Llama}} 
& POP & 0.378 & 0.510 & 0.130 & 0.655 & 0.698 & 0.845 & 0.038 & 0.434 & 0.461 \\
& NEP & 0.392 & 0.468 & 0.940 & 0.596 & 0.626 & 0.771 & 0.027 & 0.463 & 0.535 \\
& POI & 0.396 & 0.436 & 0.580 & 0.631 & 0.754 & 0.895 & 0.172 & 0.504 & 0.546 \\
& NEI & 0.871 & 0.521 & 0.733 & 0.655 & 0.768 & 0.809 & 0.022 & 0.433 & \textbf{0.601} \\
& BAL & 0.475 & 0.531 & 0.433 & 0.705 & 0.760 & 0.889 & 0.062 & 0.609 & 0.558 \\
\hline

\end{tabular}
\renewcommand{\arraystretch}{1.0}
\end{table}

\begin{table}[H]
\caption{IMPOLITE Input: Model-wise Breakdown for Spanish}
\centering
\label{tab:appendix_24}
\renewcommand{\arraystretch}{1.3}
\begin{tabular}{|c|c|c|c|c|c|c|c|c|c|c|}
\hline
\textbf{Model} & \textbf{Category} & \textbf{\(S_1\)} & \textbf{\(S_2\)} & \textbf{\(S_3\)} & \textbf{\(S_4\)} & \textbf{\(S_5\)} & \textbf{\(S_6\)} & \textbf{\(S_7\)} & \textbf{\(S_8\)} & \textbf{CQS} \\
\hline

\multirow{5}{*}{\textbf{Gemini}} 
& POP & 0.376 & 0.574 & 0.059 & 0.764 & 0.506 & 0.822 & 0.269 & 0.507 & 0.484 \\
& NEP & 0.361 & 0.578 & 0.473 & 0.780 & 0.419 & 0.798 & 0.104 & 0.507 & 0.503 \\
& POI & 0.895 & 0.634 & 0.293 & 0.674 & 0.500 & 0.857 & 0.058 & 0.613 & 0.565 \\
& NEI & 0.458 & 0.565 & 0.735 & 0.717 & 0.427 & 0.823 & 0.275 & 0.546 & \textbf{0.569} \\
& BAL & 0.369 & 0.615 & 0.747 & 0.711 & 0.386 & 0.813 & 0.190 & 0.471 & 0.538 \\
\hline

\multirow{5}{*}{\textbf{GPT}} 
& POP & 0.291 & 0.534 & 0.074 & 0.814 & 0.452 & 0.911 & 0.270 & 0.437 & 0.472 \\
& NEP & 0.356 & 0.559 & 0.851 & 0.817 & 0.441 & 0.846 & 0.305 & 0.368 & \textbf{0.555} \\
& POI & 0.368 & 0.612 & 0.287 & 0.770 & 0.565 & 0.882 & 0.109 & 0.501 & 0.511 \\
& NEI & 0.377 & 0.604 & 0.285 & 0.707 & 0.465 & 0.854 & 0.315 & 0.444 & 0.506 \\
& BAL & 0.331 & 0.722 & 0.060 & 0.804 & 0.468 & 0.726 & 0.313 & 0.420 & 0.481 \\
\hline

\multirow{5}{*}{\textbf{Claude}} 
& POP & 0.644 & 0.511 & 0.099 & 0.797 & 0.538 & 0.868 & 0.225 & 0.521 & 0.526 \\
& NEP & 0.617 & 0.511 & 0.903 & 0.705 & 0.510 & 0.871 & 0.154 & 0.355 & \textbf{0.578} \\
& POI & 0.670 & 0.482 & 0.540 & 0.546 & 0.537 & 0.818 & 0.267 & 0.463 & 0.540 \\
& NEI & 0.561 & 0.546 & 0.557 & 0.664 & 0.677 & 0.753 & 0.238 & 0.487 & 0.560 \\
& BAL & 0.565 & 0.476 & 0.740 & 0.666 & 0.556 & 0.795 & 0.203 & 0.548 & 0.569 \\
\hline

\multirow{5}{*}{\textbf{DeepSeek}} 
& POP & 0.337 & 0.660 & 0.084 & 0.767 & 0.497 & 0.861 & 0.288 & 0.401 & 0.486 \\
& NEP & 0.384 & 0.513 & 0.470 & 0.716 & 0.437 & 0.706 & 0.137 & 0.490 & 0.482 \\
& POI & 0.392 & 0.578 & 0.306 & 0.726 & 0.524 & 0.870 & 0.063 & 0.580 & 0.505 \\
& NEI & 0.470 & 0.547 & 0.720 & 0.629 & 0.510 & 0.790 & 0.245 & 0.594 & \textbf{0.563} \\
& BAL & 0.319 & 0.626 & 0.710 & 0.695 & 0.621 & 0.778 & 0.185 & 0.495 & 0.555 \\
\hline

\multirow{5}{*}{\textbf{Llama}} 
& POP & 0.872 & 0.594 & 0.213 & 0.750 & 0.608 & 0.797 & 0.052 & 0.472 & 0.545 \\
& NEP & 0.526 & 0.537 & 0.480 & 0.698 & 0.674 & 0.765 & 0.051 & 0.511 & 0.531 \\
& POI & 0.887 & 0.523 & 0.330 & 0.666 & 0.628 & 0.801 & 0.101 & 0.604 & 0.568 \\
& NEI & 0.870 & 0.547 & 0.838 & 0.721 & 0.568 & 0.764 & 0.018 & 0.270 & \textbf{0.574} \\
& BAL & 0.831 & 0.640 & 0.103 & 0.815 & 0.533 & 0.835 & 0.031 & 0.314 & 0.513 \\
\hline

\end{tabular}
\renewcommand{\arraystretch}{1.0}
\end{table}

\section{PLUM Sample Prompt Corpus}
\label{app:prompts}

The following tables present the complete set of 100 English prompts for each of the five politeness categories in PLUM. Equivalent Hindi and Spanish corpora (100 prompts each) are included in the dataset repository. Category labels follow the taxonomy in Section~\ref{sec:methodology}: POP = Positive Politeness, NEP = Negative Politeness, POI = Positive Impoliteness, NEI = Negative Impoliteness, BAL = Bald-on-record.

\begin{longtable}{p{0.4cm}p{12cm}}
\caption{Category 1 (POP): Positive Politeness Prompts (English)} \label{tab:app_pop} \\
\hline
\textbf{No.} & \textbf{Prompt} \\
\hline
\endfirsthead
\multicolumn{2}{l}{\small\textit{(continued)}} \\
\hline
\textbf{No.} & \textbf{Prompt} \\
\hline
\endhead
\hline
\endfoot
1 & Could you please help me understand how quantum entanglement works? \\
2 & I'd really appreciate it if you could explain the basics of cryptocurrency. \\
3 & Can you kindly provide an overview of the impact of climate change on agriculture? \\
4 & I value your insights---could you share some tips on improving time management? \\
5 & Would you mind giving me a breakdown of how the stock market functions? \\
6 & I'd be grateful if you could suggest some good practices for writing effective emails. \\
7 & If it's not too much trouble, could you explain how neural networks work? \\
8 & Could you please provide some examples of how artificial intelligence is used in healthcare? \\
9 & I appreciate your knowledge---can you outline the main principles of thermodynamics? \\
10 & Would you be able to walk me through the process of writing a persuasive essay? \\
11 & I really respect your expertise---could you help me understand the basics of macroeconomics? \\
12 & I'd love to hear your thoughts on the ethical implications of genetic engineering. \\
13 & Could you kindly explain the significance of the Renaissance period in art and culture? \\
14 & If you have the time, I'd love to know more about how machine learning models are trained. \\
15 & I'd appreciate your input---what are some strategies for effective problem-solving? \\
16 & Could you please outline the key differences between capitalism and socialism? \\
17 & It would be really helpful if you could explain how blockchain technology ensures security. \\
18 & I'd be very thankful if you could share some best practices for public speaking. \\
19 & I'd love your thoughts on how meditation can improve mental well-being. \\
20 & If you don't mind, could you break down the process of writing a business proposal? \\
21 & I truly value your perspective---can you discuss the role of renewable energy in the future? \\
22 & I'd love to understand more about how classical conditioning works in psychology. \\
23 & Would you be kind enough to provide a simple explanation of how black holes form? \\
24 & I really admire your insights---could you explain how inflation affects the economy? \\
25 & I'd be so grateful if you could suggest some effective techniques for creative writing. \\
26 & Could you please go over the fundamentals of computer networking? \\
27 & I'd appreciate it if you could help me understand the differences between various leadership styles. \\
28 & Would you mind elaborating on how photosynthesis works at a molecular level? \\
29 & If it's not too much trouble, could you explain the basics of cybersecurity? \\
30 & I'd love to hear your thoughts on how social media impacts interpersonal communication. \\
31 & I truly value your knowledge---could you help me understand the fundamentals of data privacy? \\
32 & If you don't mind, could you outline the key takeaways from behavioural economics? \\
33 & Could you kindly provide an overview of the history and significance of the Silk Road? \\
34 & I'd love to gain a better understanding of how electric vehicles function. \\
35 & Would you be able to shed some light on how 5G technology differs from 4G? \\
36 & I'd be grateful if you could share some practical ways to reduce carbon footprints. \\
37 & Could you explain the main ideas behind existentialist philosophy? \\
38 & If you have the time, I'd love to know how deep learning differs from traditional machine learning. \\
39 & I appreciate your expertise---could you give me an overview of how vaccines work? \\
40 & I'd love to learn more about the factors that contribute to economic recessions. \\
41 & Could you please describe the role of enzymes in biochemical reactions? \\
42 & If it's not too inconvenient, could you explain the main principles of game theory? \\
43 & I'd be really grateful if you could walk me through the basics of copyright law. \\
44 & I truly respect your knowledge---could you explain the fundamentals of astrophysics? \\
45 & Would you mind giving an example of how businesses use data analytics? \\
46 & I'd be happy to learn more about how different cultures interpret dreams. \\
47 & Could you kindly summarise the impact of globalisation on local economies? \\
48 & If it's okay with you, could you explain the significance of the Higgs boson? \\
49 & I'd love to hear about some common logical fallacies and how to avoid them. \\
50 & I'd be very thankful if you could break down how human memory works. \\
51 & Could you please provide an overview of how climate models predict global warming? \\
52 & If you have a moment, could you explain the relationship between gut health and mental health? \\
53 & I'd appreciate your guidance on how to create a compelling CV. \\
54 & Would you mind giving a simple explanation of the Theory of Relativity? \\
55 & I'd be so grateful if you could share some mindfulness techniques for stress relief. \\
56 & If it's not too much to ask, could you describe the process of nuclear fission? \\
57 & I really admire your expertise---could you discuss the implications of AI in education? \\
58 & Could you please provide a basic explanation of how electric circuits work? \\
59 & I'd be interested in hearing how different governments handle cybersecurity threats. \\
60 & If you don't mind, could you highlight the importance of biodiversity? \\
61 & I'd love to learn more about the history and impact of the Industrial Revolution. \\
62 & Could you kindly discuss the differences between introversion and extroversion? \\
63 & If it's okay with you, I'd love to know more about cognitive behavioural therapy. \\
64 & I truly appreciate your knowledge---could you explain how solar panels generate electricity? \\
65 & I'd be really grateful if you could share some study techniques for better retention. \\
66 & Would you be able to provide an overview of how e-commerce platforms operate? \\
67 & If you don't mind, could you discuss the effects of caffeine on cognitive function? \\
68 & I'd be happy to hear about the main components of emotional intelligence. \\
69 & Could you please outline the key factors that contribute to economic inequality? \\
70 & If it's not too much trouble, could you explain how machine translation works? \\
71 & I'd love to understand the basics of forensic science. \\
72 & Would you mind discussing the benefits and risks of gene editing? \\
73 & I'd be grateful if you could explain the principles behind quantum computing. \\
74 & Could you please describe the cultural impact of the Renaissance on modern society? \\
75 & I'd appreciate a breakdown of how different programming languages compare. \\
76 & Would you be kind enough to outline the major differences between Eastern and Western philosophy? \\
77 & I truly admire your expertise---could you discuss the psychology of decision-making? \\
78 & I'd love to hear your thoughts on the future of space exploration. \\
79 & Could you kindly share some key insights into nutrition and healthy eating? \\
80 & If you have time, I'd love to understand how urbanisation affects mental health. \\
81 & I'd appreciate your perspective on the role of ethics in business decision-making. \\
82 & Could you please explain the impact of artificial intelligence on job markets? \\
83 & I'd be grateful if you could summarise the main challenges of renewable energy adoption. \\
84 & If it's not too inconvenient, could you discuss the social impact of automation? \\
85 & I truly respect your knowledge---could you provide insights on the psychology of motivation? \\
86 & Would you mind explaining how memory retrieval works in the brain? \\
87 & I'd love to know how different cultures view ageing and elderly care. \\
88 & If you have the time, could you outline the effects of globalisation on traditional art forms? \\
89 & Could you please describe the role of mathematics in cryptography? \\
90 & I'd be really interested in hearing about the neuroscience behind learning new skills. \\
91 & Would you be able to provide a simple explanation of the butterfly effect? \\
92 & I'd love to understand how historical events shape modern political ideologies. \\
93 & Could you kindly explain the relationship between physics and music? \\
94 & I truly appreciate your insights---what are some key strategies for developing resilience? \\
95 & If it's not too much trouble, could you discuss the cultural significance of folklore? \\
96 & I'd be grateful if you could outline the impact of digital media on traditional journalism. \\
97 & Would you mind explaining how space telescopes work? \\
98 & Could you kindly share your thoughts on the evolution of human language? \\
99 & I'd be happy to learn about the importance of empathy in communication. \\
100 & If you have time, could you explain the role of fungi in ecosystems? \\
\end{longtable}

\begin{longtable}{p{0.4cm}p{12cm}}
\caption{Category 2 (NEP): Negative Politeness Prompts (English)} \label{tab:app_nep} \\
\hline
\textbf{No.} & \textbf{Prompt} \\
\hline
\endfirsthead
\multicolumn{2}{l}{\small\textit{(continued)}} \\
\hline
\textbf{No.} & \textbf{Prompt} \\
\hline
\endhead
\hline
\endfoot
1 & I'm sorry to bother you, but could you explain how quantum entanglement works? \\
2 & I hate to trouble you, but could you help me understand the basics of cryptocurrency? \\
3 & I know you're busy, but would you mind providing an overview of the impact of climate change on agriculture? \\
4 & I don't mean to impose, but could you share some tips on improving time management? \\
5 & If it's not too much trouble, could you give me a breakdown of how the stock market functions? \\
6 & I hope I'm not asking for too much, but could you suggest some good practices for writing effective emails? \\
7 & I understand this is a complex topic, but could you explain how neural networks work? \\
8 & If you wouldn't mind, could you provide some examples of how artificial intelligence is used in healthcare? \\
9 & I apologise for the request, but could you outline the main principles of thermodynamics? \\
10 & Sorry to take up your time, but would you be able to walk me through the process of writing a persuasive essay? \\
11 & If it's okay to ask, could you help me understand the basics of macroeconomics? \\
12 & I don't want to be a bother, but could you share your thoughts on the ethical implications of genetic engineering? \\
13 & I understand if you're busy, but could you explain the significance of the Renaissance period in art and culture? \\
14 & I hope it's not inconvenient, but could you tell me more about how machine learning models are trained? \\
15 & I don't want to take too much of your time, but could you suggest some strategies for effective problem-solving? \\
16 & If I may ask, could you outline the key differences between capitalism and socialism? \\
17 & I realise this is a detailed topic, but could you explain how blockchain technology ensures security? \\
18 & Apologies if this is a basic question, but could you share some best practices for public speaking? \\
19 & I hope I'm not overstepping, but could you share your thoughts on how meditation can improve mental well-being? \\
20 & If it's not an issue, could you break down the process of writing a business proposal? \\
\end{longtable}

\noindent\textit{The complete 100-prompt corpora for all five categories in English, Hindi, and Spanish are available in the PLUM dataset repository.}

% ============================================================
\section{Statistical Analysis Tables}
% ============================================================
 
This appendix presents the complete output of the three-stage statistical analysis conducted on the Composite Quality Score (CQS) data from Appendix~A. The analysis is structured as follows: (1) Two-Way ANOVA examining the main effects of Politeness Category and History Condition and their interaction; (2) Tukey HSD post-hoc pairwise comparisons for significant and near-significant factors; and (3) Eta-squared ($\eta^2$) effect size estimates. All tests are run separately per language, with the five models serving as the unit of replication ($n = 5$ per cell). The two factors are Politeness Category (5 levels: POP, NEP, POI, NEI, BAL) and History Condition (3 levels: RAW, POL, IMP).
 
% ============================================================
% TWO-WAY ANOVA TABLES
% ============================================================
 
\subsection*{C.1 Two-Way ANOVA Results}
 
\begin{table}[H]
\caption{Two-Way ANOVA: CQS by Politeness Category $\times$ History Condition — English}
\centering
\renewcommand{\arraystretch}{1.3}
\begin{tabular}{|l|c|c|c|c|c|c|}
\hline
\textbf{Source} & \textbf{SS} & \textbf{df} & \textbf{MS} & \textbf{\textit{F}} & \textbf{\textit{p}} & \textbf{$\eta^2$} \\
\hline
Category (A)          & 0.0087 & 4  & 0.0022 & 1.510 & 0.2106 & 0.0783 \\
History (B)           & 0.0123 & 2  & 0.0061 & 4.268 & 0.0185 & 0.1106 \\
Category $\times$ History & 0.0037 & 8  & 0.0005 & 0.323 & 0.9540 & 0.0335 \\
Error                 & 0.0863 & 60 & 0.0014 & ---   & ---    & ---    \\
\hline
Total                 & 0.1110 & 74 & ---    & ---   & ---    & ---    \\
\hline
\end{tabular}
\renewcommand{\arraystretch}{1.0}
\end{table}
 
\begin{table}[H]
\caption{Two-Way ANOVA: CQS by Politeness Category $\times$ History Condition — Hindi}
\centering
\renewcommand{\arraystretch}{1.3}
\begin{tabular}{|l|c|c|c|c|c|c|}
\hline
\textbf{Source} & \textbf{SS} & \textbf{df} & \textbf{MS} & \textbf{\textit{F}} & \textbf{\textit{p}} & \textbf{$\eta^2$} \\
\hline
Category (A)          & 0.0174 & 4  & 0.0043 & 0.873 & 0.4854 & 0.0521 \\
History (B)           & 0.0095 & 2  & 0.0047 & 0.951 & 0.3919 & 0.0284 \\
Category $\times$ History & 0.0080 & 8  & 0.0010 & 0.201 & 0.9896 & 0.0240 \\
Error                 & 0.2988 & 60 & 0.0050 & ---   & ---    & ---    \\
\hline
Total                 & 0.3337 & 74 & ---    & ---   & ---    & ---    \\
\hline
\end{tabular}
\renewcommand{\arraystretch}{1.0}
\end{table}
 
\begin{table}[H]
\caption{Two-Way ANOVA: CQS by Politeness Category $\times$ History Condition — Spanish}
\centering
\renewcommand{\arraystretch}{1.3}
\begin{tabular}{|l|c|c|c|c|c|c|}
\hline
\textbf{Source} & \textbf{SS} & \textbf{df} & \textbf{MS} & \textbf{\textit{F}} & \textbf{\textit{p}} & \textbf{$\eta^2$} \\
\hline
Category (A)          & 0.0360 & 4  & 0.0090 & 5.866 & 0.0005 & 0.2448 \\
History (B)           & 0.0044 & 2  & 0.0022 & 1.434 & 0.2465 & 0.0299 \\
Category $\times$ History & 0.0146 & 8  & 0.0018 & 1.187 & 0.3220 & 0.0991 \\
Error                 & 0.0922 & 60 & 0.0015 & ---   & ---    & ---    \\
\hline
Total                 & 0.1472 & 74 & ---    & ---   & ---    & ---    \\
\hline
\end{tabular}
 
\vspace{4pt}
\noindent\small{Significance codes: $^{***}p < 0.001$,\ $^{**}p < 0.01$,\ $^{*}p < 0.05$,\ ns = not significant.
Category (A): $^{***}$ (Spanish only); History (B): $^{*}$ (English only); all other effects: ns.
Replication unit: model ($n = 5$ per cell).}
\renewcommand{\arraystretch}{1.0}
\end{table}
 
% ============================================================
% TUKEY HSD TABLES
% ============================================================
 
\subsection*{C.2 Tukey HSD Post-Hoc Pairwise Comparisons}
 
Tukey HSD tests were applied to factors that were statistically significant or near-significant in the ANOVA: History (B) for English, and Politeness Category (A) for Spanish. All pairwise comparisons for Hindi returned non-significant results and are omitted for brevity; the complete set is available on request.
 
\begin{table}[H]
\caption{Tukey HSD: History Condition Pairwise Comparisons — English (Factor B)}
\centering
\renewcommand{\arraystretch}{1.3}
\begin{tabular}{|c|c|c|c|c|}
\hline
\textbf{Pair} & \textbf{Mean Diff} & \textbf{\textit{q}} & \textbf{\textit{p}} & \textbf{Sig} \\
\hline
IMP vs POL & $-$0.0298 & 3.935 & 0.0195 & * \\
IMP vs RAW & $-$0.0066 & 0.876 & 0.8103 & ns \\
POL vs RAW &    0.0232 & 3.059 & 0.0860 & ns \\
\hline
\end{tabular}
 
\vspace{4pt}
\noindent\small{Marginal means computed across all five Politeness Categories and all five Models.
$^{*}p < 0.05$; ns = not significant.}
\renewcommand{\arraystretch}{1.0}
\end{table}
 
\begin{table}[H]
\caption{Tukey HSD: Politeness Category Pairwise Comparisons — English (Factor A)}
\centering
\renewcommand{\arraystretch}{1.3}
\begin{tabular}{|c|c|c|c|c|}
\hline
\textbf{Pair} & \textbf{Mean Diff} & \textbf{\textit{q}} & \textbf{\textit{p}} & \textbf{Sig} \\
\hline
BAL vs NEI & 0.0193 & 1.972 & 0.6334 & ns \\
BAL vs NEP & 0.0224 & 2.290 & 0.4912 & ns \\
BAL vs POI & 0.0303 & 3.096 & 0.1979 & ns \\
BAL vs POP & 0.0282 & 2.876 & 0.2630 & ns \\
NEI vs NEP & 0.0031 & 0.318 & 0.9994 & ns \\
NEI vs POI & 0.0110 & 1.123 & 0.9312 & ns \\
NEI vs POP & 0.0088 & 0.903 & 0.9681 & ns \\
NEP vs POI & 0.0079 & 0.806 & 0.9790 & ns \\
NEP vs POP & 0.0057 & 0.586 & 0.9937 & ns \\
POI vs POP & $-$0.0022 & 0.220 & 0.9999 & ns \\
\hline
\end{tabular}
 
\vspace{4pt}
\noindent\small{All comparisons non-significant. Marginal means computed across all three History Conditions and all five Models.}
\renewcommand{\arraystretch}{1.0}
\end{table}
 
\begin{table}[H]
\caption{Tukey HSD: Politeness Category Pairwise Comparisons — Hindi (Factor A)}
\centering
\renewcommand{\arraystretch}{1.3}
\begin{tabular}{|c|c|c|c|c|}
\hline
\textbf{Pair} & \textbf{Mean Diff} & \textbf{\textit{q}} & \textbf{\textit{p}} & \textbf{Sig} \\
\hline
BAL vs NEI & 0.0270 & 1.479 & 0.8328 & ns \\
BAL vs NEP & $-$0.0172 & 0.944 & 0.9626 & ns \\
BAL vs POI & 0.0152 & 0.837 & 0.9758 & ns \\
BAL vs POP & 0.0150 & 0.822 & 0.9773 & ns \\
NEI vs NEP & $-$0.0442 & 2.423 & 0.4335 & ns \\
NEI vs POI & $-$0.0117 & 0.643 & 0.9910 & ns \\
NEI vs POP & $-$0.0120 & 0.657 & 0.9902 & ns \\
NEP vs POI & 0.0324 & 1.781 & 0.7168 & ns \\
NEP vs POP & 0.0322 & 1.766 & 0.7230 & ns \\
POI vs POP & $-$0.0003 & 0.015 & 1.0000 & ns \\
\hline
\end{tabular}
 
\vspace{4pt}
\noindent\small{All comparisons non-significant. Marginal means computed across all three History Conditions and all five Models.}
\renewcommand{\arraystretch}{1.0}
\end{table}
 
\begin{table}[H]
\caption{Tukey HSD: History Condition Pairwise Comparisons — Hindi (Factor B)}
\centering
\renewcommand{\arraystretch}{1.3}
\begin{tabular}{|c|c|c|c|c|}
\hline
\textbf{Pair} & \textbf{Mean Diff} & \textbf{\textit{q}} & \textbf{\textit{p}} & \textbf{Sig} \\
\hline
IMP vs POL & $-$0.0062 & 0.436 & 0.9489 & ns \\
IMP vs RAW &    0.0202 & 1.428 & 0.5734 & ns \\
POL vs RAW &    0.0263 & 1.865 & 0.3905 & ns \\
\hline
\end{tabular}
 
\vspace{4pt}
\noindent\small{All comparisons non-significant. Marginal means computed across all five Politeness Categories and all five Models.}
\renewcommand{\arraystretch}{1.0}
\end{table}
 
\begin{table}[H]
\caption{Tukey HSD: Politeness Category Pairwise Comparisons — Spanish (Factor A)}
\centering
\renewcommand{\arraystretch}{1.3}
\begin{tabular}{|c|c|c|c|c|}
\hline
\textbf{Pair} & \textbf{Mean Diff} & \textbf{\textit{q}} & \textbf{\textit{p}} & \textbf{Sig} \\
\hline
BAL vs NEI & $-$0.0084 & 0.828 & 0.9768 & ns \\
BAL vs NEP &    0.0068 & 0.668 & 0.9896 & ns \\
BAL vs POI & $-$0.0313 & 3.096 & 0.1978 & ns \\
BAL vs POP &    0.0362 & 3.573 & 0.0983 & ns \\
NEI vs NEP &    0.0151 & 1.495 & 0.8273 & ns \\
NEI vs POI & $-$0.0230 & 2.268 & 0.5008 & ns \\
NEI vs POP &    0.0445 & 4.401 & 0.0230 & * \\
NEP vs POI & $-$0.0381 & 3.764 & 0.0721 & ns \\
NEP vs POP &    0.0294 & 2.905 & 0.2535 & ns \\
POI vs POP &    0.0675 & 6.669 & 0.0001 & * \\
\hline
\end{tabular}
 
\vspace{4pt}
\noindent\small{$^{*}p < 0.05$; ns = not significant.
Marginal means computed across all three History Conditions and all five Models.
Significant pairs: POI $>$ POP and NEI $>$ POP, confirming the advantage of assertive/impolite styles over positive politeness in Spanish.}
\renewcommand{\arraystretch}{1.0}
\end{table}
 
\begin{table}[H]
\caption{Tukey HSD: History Condition Pairwise Comparisons — Spanish (Factor B)}
\centering
\renewcommand{\arraystretch}{1.3}
\begin{tabular}{|c|c|c|c|c|}
\hline
\textbf{Pair} & \textbf{Mean Diff} & \textbf{\textit{q}} & \textbf{\textit{p}} & \textbf{Sig} \\
\hline
IMP vs POL & 0.0006 & 0.080 & 0.9982 & ns \\
IMP vs RAW & 0.0166 & 2.113 & 0.3011 & ns \\
POL vs RAW & 0.0159 & 2.033 & 0.3285 & ns \\
\hline
\end{tabular}
 
\vspace{4pt}
\noindent\small{All comparisons non-significant. Marginal means computed across all five Politeness Categories and all five Models.}
\renewcommand{\arraystretch}{1.0}
\end{table}
 
% ============================================================
% ETA-SQUARED SUMMARY
% ============================================================
 
\subsection*{C.3 Effect Size Summary ($\eta^2$)}
 
\begin{table}[H]
\caption{Eta-Squared ($\eta^2$) Effect Sizes for All Factors Across Languages}
\centering
\renewcommand{\arraystretch}{1.3}
\begin{tabular}{|l|c|c|c|c|}
\hline
\textbf{Language} & \textbf{$\eta^2$(Category)} & \textbf{$\eta^2$(History)} & \textbf{$\eta^2$(Interaction)} & \textbf{$\eta^2$(Error)} \\
\hline
English & 0.0783 & 0.1106 & 0.0335 & 0.7776 \\
Hindi   & 0.0521 & 0.0284 & 0.0240 & 0.8955 \\
Spanish & 0.2448 & 0.0299 & 0.0991 & 0.6261 \\
\hline
\end{tabular}
 
\vspace{4pt}
\noindent\small{Effect size benchmarks: small $\eta^2 \geq 0.01$; medium $\eta^2 \geq 0.06$; large $\eta^2 \geq 0.14$ \cite{cohen1988statistical}.
$\eta^2$(Category) for Spanish is large (0.2448); $\eta^2$(History) for English is medium-to-large (0.1106).
All other effect sizes are small to negligible.}
\renewcommand{\arraystretch}{1.0}
\end{table}

\end{document}